\documentclass{article} 
\usepackage[final]{colm2025_conference}

\usepackage[utf8]{inputenc}
\usepackage{microtype}
\usepackage{hyperref}
\usepackage{url}
\usepackage{booktabs}

\usepackage{amsmath}
\usepackage{graphicx}
\usepackage{caption}
\usepackage{booktabs}
\usepackage{lineno}
\usepackage{textcomp}
\usepackage{tcolorbox}
\usepackage{varwidth}

\usepackage[T1]{fontenc}

\definecolor{darkblue}{rgb}{0, 0, 0.5}
\hypersetup{colorlinks=true, citecolor=darkblue, linkcolor=darkblue, urlcolor=darkblue}

\usepackage{enumitem}

\title{Can Large Language Models Develop Strategic Reasoning? \\ Post-training Insights from Learning Chess
}

\author{
  Dongyoon Hwang$^{1,2}$\thanks{Equal contributions. This work was done during an internship at KRAFTON.}
  \hspace{0.7em}
  Hojoon Lee$^{1,2}$\footnotemark[1]
  \hspace{0.7em}  
  Jaegul Choo$^1$
  \hspace{0.4em}
  Dongmin Park$^2$\thanks{Co-advising.}
  \hspace{0.5em}
  Jongho Park$^{2,3}$\footnotemark[2]\\
  $^1$KAIST AI\qquad
  $^2$KRAFTON\qquad
  $^3$UC Berkeley\\
  \texttt{\{godnpeter, joonleesky\}@kaist.ac.kr
  }
}

\begin{document}

\ifcolmsubmission
\linenumbers
\fi

\maketitle

\begin{abstract}
While reinforcement learning (RL) for large language models (LLMs) has shown promise in mathematical reasoning, \textit{strategic reasoning} for LLMs using RL remains largely unexplored.
We investigate whether LLMs can develop strategic reasoning capabilities through RL in chess.
To this end, we leverage a chess-pretrained action-value network to provide dense reward on the LLM's output move quality, which can be seen as a form of knowledge distillation. 
Our experiments show that our distillation-based dense rewards often outperform sparse binary rewards. However, surprisingly, all models plateau far below expert levels.
We provide SFT and RL ablations on chess reasoning training and find evidence that this limitation stems from a deficit in the pretrained models' internal understanding of chess—a deficit which RL alone may not be able to fully overcome. The code is available at \url{https://github.com/krafton-ai/Chess-R1}.
\end{abstract}

\vspace{3mm}

\section{Introduction}

Reinforcement learning with verifiable rewards (RLVR) has shown strong performance in developing mathematical reasoning capabilities for large language models (LLMs)~\citep{guo2025deepseek,li2025llms,yu2025dapo}.
While these successes highlight LLMs' capacity for logical thinking, a critical dimension of intelligence remains largely unexplored: \textbf{strategic reasoning}—the ability to plan, anticipate adversary actions, and make decisions in multi-agent environments. Beyond logical reasoning in static settings, strategic reasoning aligns more with real-world scenarios such as games, negotiation, and market competitions~\citep{zhang2024llm, park2025orak}. 

To investigate this gap, we turn to \textit{chess}, a strategic game demanding deep strategic reasoning abilities such as positional evaluation, long-term planning, and reasoning about an opponent's intentions. 
In addition, chess offers a favorable environment for applying RLVR on LLMs, as it provides abundant publicly available game records and human-annotated reasoning about optimal moves. 
Given such a testbed for examining strategic reasoning, we raise the following research question:

\begin{center}
\textit{Can LLMs develop strategic reasoning capabilities through RLVR with chess?}
\end{center}

To this end, we train Qwen2.5~\citep{qwen2025qwen25technicalreport} and Llama3.1~\citep{grattafiori2024llama3} models to predict the next best move in chess using GRPO~\citep{shao2024deepseekmath}.
Unlike typical RLVR approaches that rely on sparse binary rewards (correct/incorrect), chess allows for dense reward signals: we leverage the fact that we can evaluate each move based on its estimated win probability, providing graded feedback proportional to the quality of the move. 
We implement this using a pre-trained chess expert as a reward model—a form of \emph{knowledge distillation} from Q-value network to LLM—which evaluates position strength and provides dense continuous rewards rather than sparse binary ones.


We conduct systematic experiments on a chess puzzle dataset, and observe the following:
\begin{itemize}
    \item Distillation-based dense rewards substantially outperforms sparse binary rewards in chess puzzle accuracy.
    \item RL fine-tuning performs stronger than supervised fine-tuning (SFT); yet, its performance plateaus at 25-30$\%$ puzzle accuracy—well below what is considered human expert performance (60–80$\%$).
    \item Additional reasoning distillation shows no RLVR gains, indicating the models fail to develop meaningful strategic reasoning despite further post-training.
\end{itemize}

Through additional failure analysis, we find that base LLMs often struggle to grasp fundamental chess rules. Thus, we hypothesize that, in contrast to logical reasoning in math domains, the limited emergence of strategic reasoning in chess may stem from insufficient exposure to chess-specific knowledge during LLM pre-training. Our empirical findings also support recent claims that RL mainly amplifies existing capabilities of pre-trained LLMs~\citep{li2025llms, zhao2025echo}, and offer insight for practitioners aiming to elicit reasoning abilities in new environments: pre-trained domain knowledge is essential to develop advanced reasoning.

\section{Post-training LLM on Chess}
\label{section:chess_training_pipeline}

\begin{figure}
    \centering
    \includegraphics[width=0.85\linewidth]{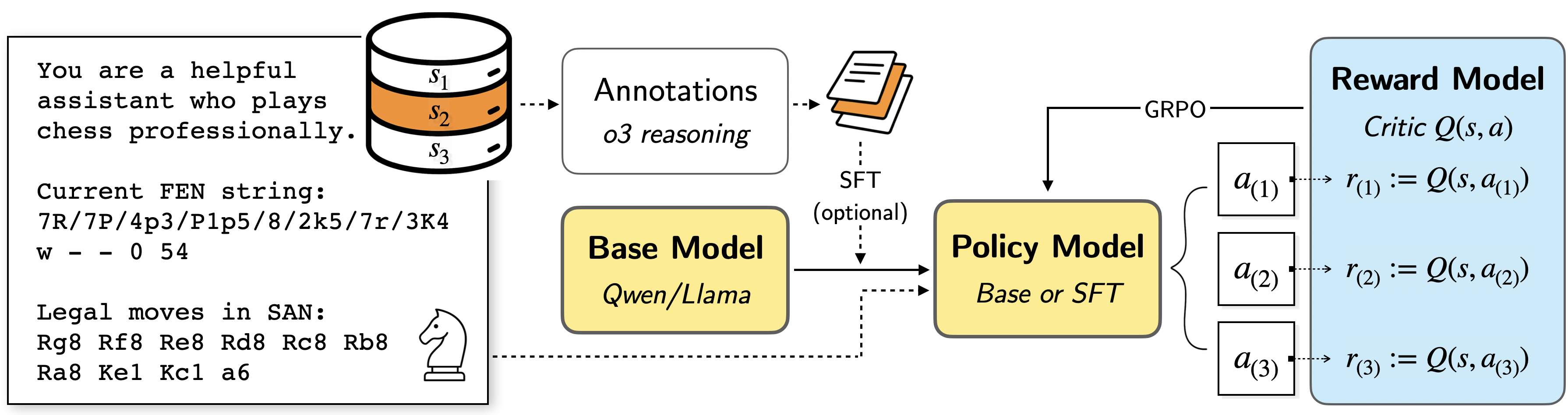}
    
    \hspace{0.5cm}{\small (a) Chess Training Example.}
    \hspace{1.5cm} {\small (b) Post-training with GRPO.} \hspace{2.7cm} 
    \caption{\textbf{Overview}: (a) A data sample from the Lichess puzzle dataset is formatted into a prompt that includes the current board state and the set of legal moves. 
     (b) At each GRPO step, the policy model generates multiple rollouts of predicted actions. A reward model evaluates these rollouts with dense feedback, including sub-optimal actions. Optionally, reasoning traces from OpenAI \texttt{o3} are fine-tuned into the base model before RL.}
    \label{fig:enter-label}
\end{figure}

\subsection{Chess Dataset}
\label{sub_section:training_data}

\paragraph{Training Data.} We use Lichess puzzles collected from the \emph{Lichess Puzzle Database}\footnote{https://database.lichess.org} as our training dataset.
Each puzzle contains an initial board position along with a list of sequential moves which, when played in order, leads to a tactical win.
To create our training dataset, we decompose each puzzle's solution trajectory into individual position-action pairs. Formally, each puzzle is denoted as a trajectory $g^{(i)} = (s^{(i)}_0, a^{*(i)}_0, s^{(i)}_1, a^{*(i)}_1, \ldots, s^{(i)}_{T_i})$, where $i \in \{1, \ldots, N\}$ indexes the $i$-th puzzle, $s^{(i)}_t$ is the board position at time step $t$, $a^{*(i)}_t$ is the optimal move, and $T_i$ is the trajectory length. We construct our position-diverse training dataset by aggregating all such pairs:
$\mathcal{D} = \{\, (s^{(i)}_t,\, a^{*(i)}_t)\mid i \in \{1,\dots,N\},\; 0 \le t < T_i \}\!$, 
resulting in a total of 19.2k training samples for our dataset $\mathcal{D}$.

\paragraph{Prompt template.}
We require a textual interface which LLMs can reason about chess positions. Based on our collected chess dataset $\mathcal{D}$, we adopt a concise prompt format using Forsyth–Edwards Notation (FEN)~\citep{edwards1994standard} format for board states and Standard Algebraic Notation (SAN) format for moves.
Alternative representations of chess data, such as full move histories in Portable Game Notation (PGN) or Universal Chess Interface (UCI) move notation, did not yield measurable improvements.
We also require the model to place reasoning in \texttt{<think>} tags and answers in \texttt{<answer>} tags~\citep{guo2025deepseek, zhao2025absoluterl}.
See Appendix~\ref{appendix:input_prompt_format},~\ref{appendix:prompt_ablations} for qualitative examples and ablation studies on prompt formatting.

\subsection{Reinforcement Learning Fine Tuning}
\label{sec:rl_finetune}

To train LLMs with RL on chess, we adopt GRPO~\citep{shao2024deepseekmath} for policy improvement, which has recently been shown to be effective by contrasting multiple rollouts of the same prompt~\citep{guo2025deepseek}.
Specifically, we employ two alternative reward schemes:
\begin{itemize}[leftmargin=10pt,noitemsep]
  \item \textbf{Sparse reward:}  
        Binary indicator whether predicted move \(\hat{a}_t\) matches optimal answer
        \(a_t^\star\):
        \begin{equation*}
          r_{\text{sparse}}
          \;=\;
          \mathbf{1}[\,\hat{a}_t = a_t^{\star}\,].
          \label{eq:r_sparse}
        \end{equation*}

  \item \textbf{Dense reward:}  
        A dense real-valued score provided by a pre-trained action-value network \(Q_{\theta}(s,a)\) from \citet{ruoss2024amortized} that predicts post-move win probability:
        \begin{equation*}
          r_{\text{dense}}
          \;=\;
          Q_{\theta}(s_t,a_t), 
          \qquad
          Q_{\theta}(s,a)\in[0,1].
          \label{eq:r_dense}
        \end{equation*}
        The critic $Q_\theta(s,a)$ is a 270\,M-parameter, 16-layer decoder-only Transformer~\citep{vaswani2017attention} trained for 10\,M optimization steps on 15B Stockfish annotated state-action pairs with the HL-Gauss loss~\citep{imani2018hlgauss}.  
        Querying this expert network during RL fine-tuning can be viewed as a form of \emph{knowledge distillation}~\citep{zhang2025distill}: dense win-rate evaluations inject the teacher’s strategic insight, guiding the student for not only optimal moves, but sub-optimal moves also.
        The critic itself attains 2299 ELO and achieves 95.4\% accuracy on Lichess puzzles.
\end{itemize}

We also include two auxiliary binary rewards: (i) \(r_{\text{fmt}}\in\{0,1\}\) for proper tag formatting (\texttt{<think>} for reasoning and \texttt{<answer>} for answers), and (ii) \(r_{\text{lang}}\in\{0,1\}\) to enforce that the output is in English.
The total training reward at step \(t\) is therefore $r_t = \lambda_{\text{sparse}} r_{\text{sparse}} + \lambda_{\text{dense}} r_{\text{dense}} + \lambda_{\text{fmt}} r_{\text{fmt}} + \lambda_{\text{lang}} r_{\text{lang}}$.
We set $(\lambda_{\text{sparse}}=1, \lambda_{\text{dense}}=0)$ for the sparse reward setting and $(\lambda_{\text{sparse}}=0, \lambda_{\text{dense}}=1)$ for the dense reward setting.

\section{Experiments}

We now present a series of experiments designed to evaluate
post-training LLM on chess with RLVR.
Further details regarding experiment setup are provided in Appendix~\ref{appendix:experiment_setup}.

\subsection{RL Fine-tuning}
\label{subsection:rl_finetuning}

\begin{figure}[h]
    \centering
    \includegraphics[width=0.96\linewidth]{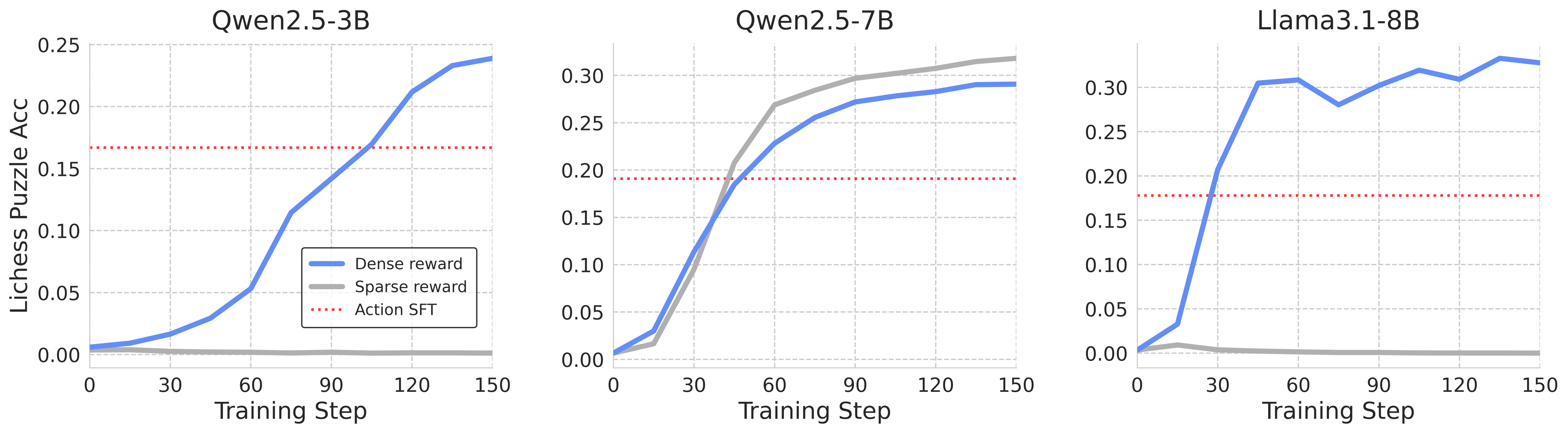}
    \caption{Evaluation performance comparison of RL fine-tuned models.}
    \label{fig:rl_learning_curve}
\end{figure}

We fine-tuned Qwen2.5-3B, Qwen2.5-7B and Llama3.1-8B on our Lichess position-action pair dataset $\mathcal{D}$.
For each model, we trained three variants: one with sparse rewards, one with dense expert guidance, and one baseline variant fine-tuned directly via supervised learning (SFT) on optimal actions, without RL. The supervised baseline helps evaluate performance when no explicit reasoning is involved.

Figure~\ref{fig:rl_learning_curve} shows that dense reward models generally outperform sparse reward variants, with sparse rewards completely failing for Qwen2.5-3B and Llama3.1-8B. Dense reward models also surpass SFT baselines, demonstrating the value of reasoning over direct action prediction.
However, all models plateau around 25-30\% puzzle accuracy, well below expert performance (1800 ELO models achieve 66.5\%~\citep{ruoss2024amortized}).
Qualitative analysis (Appendix~\ref{appendix:qualitative_results}) reveals that while models show structured reasoning and systematic move exploration, they lack strategic coherence and long-term evaluation.
In summary, although RL improves tactical reasoning particularly with expert-guided feedback, achieving expert-level strategic chess understanding from scalar rewards remains challenging for LLMs.

\begin{figure}[t]
    \centering
    \begin{minipage}[t]{0.58\linewidth}
        \centering
        \includegraphics[width=\linewidth]{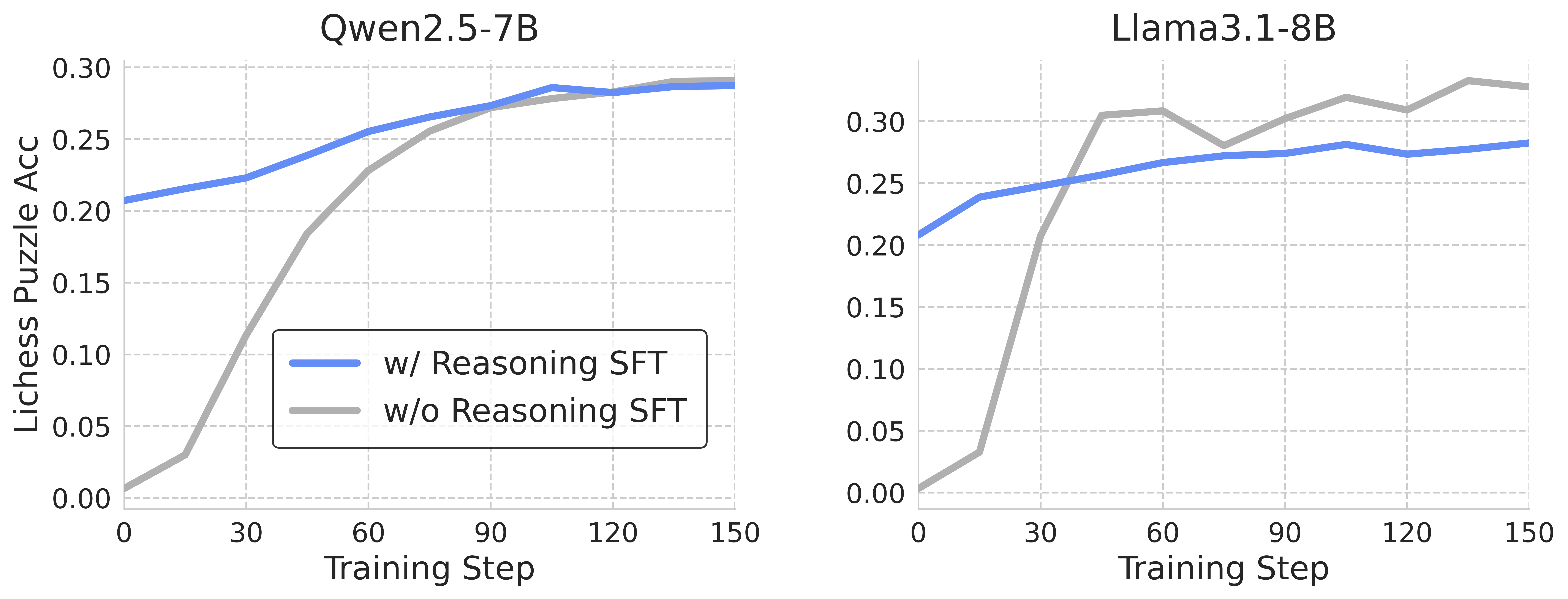}
        \vspace{-0.65cm}
        \caption{Evaluation performance of models trained with reasoning SFT followed by RL fine-tuning.}
        \label{fig:o3_rl_learning_curve}
    \end{minipage}
    \hfill
    \begin{minipage}[t]{0.4\linewidth}
        \centering
        \vspace{-3.0cm}  
        \scriptsize 
        \setlength{\tabcolsep}{2pt}
        \renewcommand{\arraystretch}{1.2}
        \begin{tabular}{lcc}
            \toprule
            \textbf{Model} & \textbf{Board State Acc. ↑} & \textbf{MATE Acc. ↑} \\
            \midrule
            Qwen2.5-3B & 0.0\% & 35.8\% \\
            Qwen2.5-3B-It & 0.0\% & 53.7\% \\
            Qwen2.5-7B & 0.0\% & 42.7\% \\
            Qwen2.5-7B-It & 0.0\% & 52.0\% \\
            Llama3.1-8B & 0.0\% & 12.7\% \\
            Llama3.1-8B-It & 0.0\% & 52.4\% \\
            \bottomrule
        \end{tabular}
        \captionof{table}{LLM performance on chess knowledge diagnostic tasks}
        \label{table:failure_analysis}
    \end{minipage}
\vspace{-0.1cm}
\end{figure}

\subsection{Reasoning SFT}

To probe the cause of the RL plateau, we asked whether domain-specific reasoning traces from advanced reasoning models could enable LLMs better leverage RL signals for chess skill acquisition and overcome the previously observed performance ceiling.
To do so, we curated a \emph{reasoning SFT} corpus of 1k high-quality reasoning traces generated by OpenAI o3, featuring candidate move evaluation, tactical assessment, and strategic justifications (See Appendix~\ref{appendix:o3_sft_reasoning_data} for details).
After performing SFT on Qwen2.5-7B and Llama3.1-8B with this corpus, we applied our GRPO RL pipeline with expert critic $Q_\theta(s,a)$ feedback.

While LLMs produced markedly more comprehensive reasoning traces after reasoning SFT, Figure~\ref{fig:o3_rl_learning_curve} shows that they disappointingly exhibited similar puzzle accuracy plateaus when subsequently trained with RL. In fact, Llama3.1-8B's performance actually declined relative to its non-SFT baseline.
These results raise questions about whether the reason behind the chess performance plateau stems not from inadequate reasoning abilities, but from insufficient chess knowledge.
Without extensive domain knowledge from pre-training, RL alone may not provide the chess-specific understanding that strategic play demands.

\subsection{Failure Analysis}
\label{subsection:failure_analysis}

To study whether the performance limitation of RLVR for chess stems from inadequate chess domain knowledge, we evaluate LLMs on two diagnostic tasks that require basic and essential but non-trivial understanding of chess rules.

\textbf{Board–state comprehension:} Given a FEN string and a legal sequence of one to five SAN moves, the model predicts the resulting position in FEN. The model must have a faithful internal simulator of chess rules to be successful in this task.

\textbf{MATE puzzle:} Each instance in the MATE dataset~\citep{wang2024MATE} presents a mid-game position together with two candidate moves. The model must identify the superior move through tactical evaluation.

As summarized in Table~\ref{table:failure_analysis}, all tested LLMs show poor performance on both diagnostics, confirming inadequate internal models of chess state transitions and tactics.
We tested both Base and Instruct variants—Instruct models for better formatting compliance, though they share the same underlying knowledge as their Base counterparts.
This shortfall strongly supports our earlier hypothesis: current models lack fundamental understanding of chess, preventing them from developing high performance through reward optimization alone.
Since models cannot reliably track game states or recognize elementary tactics, reward optimization fails to develop high-quality strategic reasoning and elicit expert-level play.

\section{Related Work}
\label{appendix:related_work}

\paragraph{RLVR for LLM reasoning.}
RLVR~\citep{lambert2024t} has emerged as a powerful paradigm to elicit LLM reasoning for tasks with deterministic evaluation criteria. 
Notably, in mathematical domains, simple rule-based rewards have proven surprisingly effective without requiring complex reward models~\citep{guo2025deepseek, yu2025dapo, wang2025reinforcement}, achieving performance comparable to, or even surpassing, that of humans on mathematical olympiads such as AIME.
Furthermore, some works have extended the application of RLVR to other domains as well~\citep{liu2025visual, zhang2025med, gurung2025learning}.

Despite these successes, fundamental questions about RLVR's success remain underexplored.
Recent studies suggest that its success often originates from abundant knowledge already present in base models rather than being genuinely learned through RL~\citep{liu2025understanding, zhao2025echo, li2025llms, shao2025spurious, yue2025does}. 
A systematic investigation across models, tasks, RL algorithms, and scale remains necessary to fully understand LLM capabilities and its limitations in developing genuine reasoning abilities through RL.

\paragraph{LLMs in chess.}
Recent work explores how LLMs can tackle chess tasks, complementing traditional chess engines.
While specialized networks like AlphaZero~\citep{silver2017alphazero} achieve superhuman performance via self-play and search, they lack human-readable insight.
Conversely, LLMs can explain moves in natural language but often lack precise strategic understanding, leading to hallucinations or illegal moves.
To bridge this gap, researchers have integrated chess engines with LLMs for fluent move commentary~\citep{kim2025commentary_chess} and developed chess-centric models like ChessGPT~\citep{feng2023chessgpt} through domain-specific pre-training. 
Fine-tuning on expert rationales from the MATE dataset~\citep{wang2024MATE} enables Llama models to outperform GPT-4 and Claude at binary move selection.
Most notably, ChessLLM~\citep{zhang2025chessllm} demonstrates that an LLM fine-tuned on 20B tokens of complete games can play chess at an expert level, attaining an Elo of 1788 without external search; however FEN-centric training risks overfitting and degrading natural language responses into fragmented notations.

Despite these advances, existing approaches primarily rely on supervised learning from human games or engine guidance rather than autonomous strategic development. Whether LLMs can develop genuine chess reasoning through RL remains largely unexplored, particularly in understanding if such learning produces transferable strategic insights or merely pattern memorization. Our work addresses this gap using chess as a controlled testbed for studying autonomous strategic development in LLMs.

\section{Conclusion \& Limitations}

We investigated whether LLMs can develop strategic reasoning through RLVR on chess, introducing a novel approach that utilizes dense rewards from pretrained action-value networks. While dense rewards often outperform sparse ones, all models plateau well below human experts, revealing fundamental limitations in current post-training approaches.
Surprisingly, despite producing more structured reasoning when trained on advanced traces from OpenAI \texttt{o3}, models still yield similar performance plateaus when subsequently trained with RL.
Our failure analysis reveals a potential root cause: current LLMs demonstrate inadequate internal chess knowledge.
These results raise speculation that RL cannot overcome impoverished domain knowledge.
While RL excels at optimizing behavior toward rewards, LLMs cannot learn \textit{de novo} the foundational knowledge necessary for strategic thinking when absent from pretraining. We hypothesize that prior RLVR successes in math domains stem from rich pretraining exposure, unlike chess which receives minimal coverage.
Our findings align with recent work showing that RL only amplifies existing capabilities~\citep{li2025llms, liu2025understanding, yue2025does}.
For strategic reasoning in new domains, adequate domain knowledge during pretraining is essential—post-training RL alone is insufficient.

\bibliography{colm2025_conference}

\begin{thebibliography}{30}
\providecommand{\natexlab}[1]{#1}
\providecommand{\url}[1]{\texttt{#1}}
\expandafter\ifx\csname urlstyle\endcsname\relax
  \providecommand{\doi}[1]{doi: #1}\else
  \providecommand{\doi}{doi: \begingroup \urlstyle{rm}\Url}\fi

\bibitem[Edwards(1994)]{edwards1994standard}
Steven~J. Edwards.
\newblock Standard: Portable game notation specification and implementation guide, 1994.
\newblock URL \url{https://ia802908.us.archive.org/26/items/pgn-standard-1994-03-12/PGN_standard_1994-03-12.txt}.

\bibitem[Feng et~al.(2023)Feng, Luo, Wang, Tang, Yang, Shao, Mguni, Du, and Wang]{feng2023chessgpt}
Xidong Feng, Yicheng Luo, Ziyan Wang, Hongrui Tang, Mengyue Yang, Kun Shao, David Mguni, Yali Du, and Jun Wang.
\newblock Chessgpt: Bridging policy learning and language modeling.
\newblock \emph{Advances in Neural Information Processing Systems}, 36:\penalty0 7216--7262, 2023.

\bibitem[Grattafiori et~al.(2024)Grattafiori, Dubey, Jauhri, Pandey, Kadian, Al-Dahle, Letman, Mathur, Schelten, Vaughan, et~al.]{grattafiori2024llama3}
Aaron Grattafiori, Abhimanyu Dubey, Abhinav Jauhri, Abhinav Pandey, Abhishek Kadian, Ahmad Al-Dahle, Aiesha Letman, Akhil Mathur, Alan Schelten, Alex Vaughan, et~al.
\newblock The llama 3 herd of models.
\newblock \emph{arXiv preprint arXiv:2407.21783}, 2024.

\bibitem[Guo et~al.(2025)Guo, Yang, Zhang, Song, Zhang, Xu, Zhu, Ma, Wang, Bi, et~al.]{guo2025deepseek}
Daya Guo, Dejian Yang, Haowei Zhang, Junxiao Song, Ruoyu Zhang, Runxin Xu, Qihao Zhu, Shirong Ma, Peiyi Wang, Xiao Bi, et~al.
\newblock Deepseek-r1: Incentivizing reasoning capability in llms via reinforcement learning.
\newblock \emph{arXiv preprint arXiv:2501.12948}, 2025.

\bibitem[Gurung \& Lapata(2025)Gurung and Lapata]{gurung2025learning}
Alexander Gurung and Mirella Lapata.
\newblock Learning to reason for long-form story generation.
\newblock \emph{arXiv preprint arXiv:2503.22828}, 2025.

\bibitem[Imani \& White(2018)Imani and White]{imani2018hlgauss}
Ehsan Imani and Martha White.
\newblock Improving regression performance with distributional losses.
\newblock In \emph{International conference on machine learning}, pp.\  2157--2166. PMLR, 2018.

\bibitem[Kim et~al.(2025)Kim, Goh, Hwang, Cho, and Ok]{kim2025commentary_chess}
Jaechang Kim, Jinmin Goh, Inseok Hwang, Jaewoong Cho, and Jungseul Ok.
\newblock Bridging the gap between expert and language models: Concept-guided chess commentary generation and evaluation.
\newblock In \emph{Proceedings of the 2025 Conference of the Nations of the Americas Chapter of the Association for Computational Linguistics: Human Language Technologies (Volume 1: Long Papers)}, Albuquerque, New Mexico, April 2025. Association for Computational Linguistics.

\bibitem[Lambert et~al.(2024)Lambert, Morrison, Pyatkin, Huang, Ivison, Brahman, Miranda, Liu, Dziri, Lyu, et~al.]{lambert2024t}
Nathan Lambert, Jacob Morrison, Valentina Pyatkin, Shengyi Huang, Hamish Ivison, Faeze Brahman, Lester James~V Miranda, Alisa Liu, Nouha Dziri, Shane Lyu, et~al.
\newblock T$\backslash$" ulu 3: Pushing frontiers in open language model post-training.
\newblock \emph{arXiv preprint arXiv:2411.15124}, 2024.

\bibitem[Li et~al.(2025)Li, Cao, Griggs, Liu, Mo, Tang, Hegde, Hakhamaneshi, Patil, Zaharia, et~al.]{li2025llms}
Dacheng Li, Shiyi Cao, Tyler Griggs, Shu Liu, Xiangxi Mo, Eric Tang, Sumanth Hegde, Kourosh Hakhamaneshi, Shishir~G Patil, Matei Zaharia, et~al.
\newblock Llms can easily learn to reason from demonstrations structure, not content, is what matters!
\newblock \emph{arXiv preprint arXiv:2502.07374}, 2025.

\bibitem[Liu et~al.(2025{\natexlab{a}})Liu, Chen, Li, Qi, Pang, Du, Lee, and Lin]{liu2025understanding}
Zichen Liu, Changyu Chen, Wenjun Li, Penghui Qi, Tianyu Pang, Chao Du, Wee~Sun Lee, and Min Lin.
\newblock Understanding r1-zero-like training: A critical perspective.
\newblock \emph{arXiv preprint arXiv:2503.20783}, 2025{\natexlab{a}}.

\bibitem[Liu et~al.(2025{\natexlab{b}})Liu, Sun, Zang, Dong, Cao, Duan, Lin, and Wang]{liu2025visual}
Ziyu Liu, Zeyi Sun, Yuhang Zang, Xiaoyi Dong, Yuhang Cao, Haodong Duan, Dahua Lin, and Jiaqi Wang.
\newblock Visual-rft: Visual reinforcement fine-tuning.
\newblock \emph{arXiv preprint arXiv:2503.01785}, 2025{\natexlab{b}}.

\bibitem[Park et~al.(2025)Park, Kim, Choi, Kim, Lee, Lee, Park, Lee, Hwang, Ahn, et~al.]{park2025orak}
Dongmin Park, Minkyu Kim, Beongjun Choi, Junhyuck Kim, Keon Lee, Jonghyun Lee, Inkyu Park, Byeong-Uk Lee, Jaeyoung Hwang, Jaewoo Ahn, et~al.
\newblock Orak: A foundational benchmark for training and evaluating llm agents on diverse video games.
\newblock \emph{arXiv preprint arXiv:2506.03610}, 2025.

\bibitem[Qwen: et~al.(2025)Qwen:, Yang, Yang, Zhang, Hui, Zheng, Yu, Li, Liu, Huang, Wei, Lin, Yang, Tu, Zhang, Yang, Yang, Zhou, Lin, Dang, Lu, Bao, Yang, Yu, Li, Xue, Zhang, Zhu, Men, Lin, Li, Tang, Xia, Ren, Ren, Fan, Su, Zhang, Wan, Liu, Cui, Zhang, and Qiu]{qwen2025qwen25technicalreport}
Qwen:, An~Yang, Baosong Yang, Beichen Zhang, Binyuan Hui, Bo~Zheng, Bowen Yu, Chengyuan Li, Dayiheng Liu, Fei Huang, Haoran Wei, Huan Lin, Jian Yang, Jianhong Tu, Jianwei Zhang, Jianxin Yang, Jiaxi Yang, Jingren Zhou, Junyang Lin, Kai Dang, Keming Lu, Keqin Bao, Kexin Yang, Le~Yu, Mei Li, Mingfeng Xue, Pei Zhang, Qin Zhu, Rui Men, Runji Lin, Tianhao Li, Tianyi Tang, Tingyu Xia, Xingzhang Ren, Xuancheng Ren, Yang Fan, Yang Su, Yichang Zhang, Yu~Wan, Yuqiong Liu, Zeyu Cui, Zhenru Zhang, and Zihan Qiu.
\newblock Qwen2.5 technical report.
\newblock 2025.
\newblock URL \url{https://arxiv.org/abs/2412.15115}.

\bibitem[Ruoss et~al.(2024)Ruoss, Del{\'e}tang, Medapati, Grau-Moya, Wenliang, Catt, Reid, Lewis, Veness, and Genewein]{ruoss2024amortized}
Anian Ruoss, Gr{\'e}goire Del{\'e}tang, Sourabh Medapati, Jordi Grau-Moya, Li~K Wenliang, Elliot Catt, John Reid, Cannada~A Lewis, Joel Veness, and Tim Genewein.
\newblock Amortized planning with large-scale transformers: A case study on chess.
\newblock \emph{Advances in Neural Information Processing Systems}, 37:\penalty0 65765--65790, 2024.

\bibitem[Shao et~al.(2025)Shao, Li, Xin, Geng, Wang, Oh, Du, Lambert, Min, Krishna, et~al.]{shao2025spurious}
Rulin Shao, Shuyue~Stella Li, Rui Xin, Scott Geng, Yiping Wang, Sewoong Oh, Simon~Shaolei Du, Nathan Lambert, Sewon Min, Ranjay Krishna, et~al.
\newblock Spurious rewards: Rethinking training signals in rlvr.
\newblock \emph{arXiv preprint arXiv:2506.10947}, 2025.

\bibitem[Shao et~al.(2024)Shao, Wang, Zhu, Xu, Song, Bi, Zhang, Zhang, Li, Wu, et~al.]{shao2024deepseekmath}
Zhihong Shao, Peiyi Wang, Qihao Zhu, Runxin Xu, Junxiao Song, Xiao Bi, Haowei Zhang, Mingchuan Zhang, YK~Li, Y~Wu, et~al.
\newblock Deepseekmath: Pushing the limits of mathematical reasoning in open language models.
\newblock \emph{arXiv preprint arXiv:2402.03300}, 2024.

\bibitem[Sheng et~al.(2024)Sheng, Zhang, Ye, Wu, Zhang, Zhang, Peng, Lin, and Wu]{sheng2024verl}
Guangming Sheng, Chi Zhang, Zilingfeng Ye, Xibin Wu, Wang Zhang, Ru~Zhang, Yanghua Peng, Haibin Lin, and Chuan Wu.
\newblock Hybridflow: A flexible and efficient rlhf framework.
\newblock \emph{arXiv preprint arXiv: 2409.19256}, 2024.

\bibitem[Silver et~al.(2017)Silver, Hubert, Schrittwieser, Antonoglou, Lai, Guez, Lanctot, Sifre, Kumaran, Graepel, et~al.]{silver2017alphazero}
David Silver, Thomas Hubert, Julian Schrittwieser, Ioannis Antonoglou, Matthew Lai, Arthur Guez, Marc Lanctot, Laurent Sifre, Dharshan Kumaran, Thore Graepel, et~al.
\newblock Mastering chess and shogi by self-play with a general reinforcement learning algorithm.
\newblock \emph{arXiv preprint arXiv:1712.01815}, 2017.

\bibitem[Vaswani et~al.(2017)Vaswani, Shazeer, Parmar, Uszkoreit, Jones, Gomez, Kaiser, and Polosukhin]{vaswani2017attention}
Ashish Vaswani, Noam Shazeer, Niki Parmar, Jakob Uszkoreit, Llion Jones, Aidan~N Gomez, {\L}ukasz Kaiser, and Illia Polosukhin.
\newblock Attention is all you need.
\newblock \emph{Advances in neural information processing systems}, 30, 2017.

\bibitem[Wang et~al.(2024)Wang, Ji, Wang, Zhao, Liu, Hou, and Wu]{wang2024MATE}
Shu Wang, Lei Ji, Renxi Wang, Wenxiao Zhao, Haokun Liu, Yifan Hou, and Ying~Nian Wu.
\newblock Explore the reasoning capability of llms in the chess testbed.
\newblock \emph{arXiv preprint arXiv:2411.06655}, 2024.

\bibitem[Wang et~al.(2025)Wang, Yang, Zeng, Ren, Liu, Peng, Cheng, He, Wang, Gao, et~al.]{wang2025reinforcement}
Yiping Wang, Qing Yang, Zhiyuan Zeng, Liliang Ren, Liyuan Liu, Baolin Peng, Hao Cheng, Xuehai He, Kuan Wang, Jianfeng Gao, et~al.
\newblock Reinforcement learning for reasoning in large language models with one training example.
\newblock \emph{arXiv preprint arXiv:2504.20571}, 2025.

\bibitem[Yu et~al.(2025)Yu, Zhang, Zhu, Yuan, Zuo, Yue, Fan, Liu, Liu, Liu, et~al.]{yu2025dapo}
Qiying Yu, Zheng Zhang, Ruofei Zhu, Yufeng Yuan, Xiaochen Zuo, Yu~Yue, Tiantian Fan, Gaohong Liu, Lingjun Liu, Xin Liu, et~al.
\newblock Dapo: An open-source llm reinforcement learning system at scale.
\newblock \emph{arXiv preprint arXiv:2503.14476}, 2025.

\bibitem[Yue et~al.(2025)Yue, Chen, Lu, Zhao, Wang, Song, and Huang]{yue2025does}
Yang Yue, Zhiqi Chen, Rui Lu, Andrew Zhao, Zhaokai Wang, Shiji Song, and Gao Huang.
\newblock Does reinforcement learning really incentivize reasoning capacity in llms beyond the base model?
\newblock \emph{arXiv preprint arXiv:2504.13837}, 2025.

\bibitem[Zhang et~al.(2025{\natexlab{a}})Zhang, Liu, Qin, Naumann, and Poon]{zhang2025med}
Sheng Zhang, Qianchu Liu, Guanghui Qin, Tristan Naumann, and Hoifung Poon.
\newblock Med-rlvr: Emerging medical reasoning from a 3b base model via reinforcement learning.
\newblock \emph{arXiv preprint arXiv:2502.19655}, 2025{\natexlab{a}}.

\bibitem[Zhang et~al.(2024)Zhang, Mao, Ge, Wang, de~Wynter, Xia, Wu, Song, Lan, and Wei]{zhang2024llm}
Yadong Zhang, Shaoguang Mao, Tao Ge, Xun Wang, Adrian de~Wynter, Yan Xia, Wenshan Wu, Ting Song, Man Lan, and Furu Wei.
\newblock Llm as a mastermind: A survey of strategic reasoning with large language models.
\newblock \emph{arXiv preprint arXiv:2404.01230}, 2024.

\bibitem[Zhang et~al.(2025{\natexlab{b}})Zhang, Han, Li, Chen, and Lin]{zhang2025chessllm}
Yinqi Zhang, Xintian Han, Haolong Li, Kedi Chen, and Shaohui Lin.
\newblock Complete chess games enable {LLM} become a chess master.
\newblock Association for Computational Linguistics, 2025{\natexlab{b}}.

\bibitem[Zhang et~al.(2025{\natexlab{c}})Zhang, Wang, Fang, Du, Huang, Wang, Lin, Pechenizkiy, Zhang, Rajmohan, et~al.]{zhang2025distill}
Yudi Zhang, Lu~Wang, Meng Fang, Yali Du, Chenghua Huang, Jun Wang, Qingwei Lin, Mykola Pechenizkiy, Dongmei Zhang, Saravan Rajmohan, et~al.
\newblock Distill not only data but also rewards: Can smaller language models surpass larger ones?
\newblock \emph{arXiv preprint arXiv:2502.19557}, 2025{\natexlab{c}}.

\bibitem[Zhao et~al.(2025{\natexlab{a}})Zhao, Wu, Yue, Wu, Xu, Lin, Wang, Wu, Zheng, and Huang]{zhao2025absoluterl}
Andrew Zhao, Yiran Wu, Yang Yue, Tong Wu, Quentin Xu, Matthieu Lin, Shenzhi Wang, Qingyun Wu, Zilong Zheng, and Gao Huang.
\newblock Absolute zero: Reinforced self-play reasoning with zero data.
\newblock \emph{arXiv preprint arXiv:2505.03335}, 2025{\natexlab{a}}.

\bibitem[Zhao et~al.(2025{\natexlab{b}})Zhao, Meterez, Kakade, Pehlevan, Jelassi, and Malach]{zhao2025echo}
Rosie Zhao, Alexandru Meterez, Sham Kakade, Cengiz Pehlevan, Samy Jelassi, and Eran Malach.
\newblock Echo chamber: Rl post-training amplifies behaviors learned in pretraining.
\newblock \emph{arXiv preprint arXiv:2504.07912}, 2025{\natexlab{b}}.

\bibitem[Zheng et~al.(2024)Zheng, Zhang, Zhang, Ye, Luo, Feng, and Ma]{zheng2024llamafactory}
Yaowei Zheng, Richong Zhang, Junhao Zhang, Yanhan Ye, Zheyan Luo, Zhangchi Feng, and Yongqiang Ma.
\newblock Llamafactory: Unified efficient fine-tuning of 100+ language models.
\newblock In \emph{Proceedings of the 62nd Annual Meeting of the Association for Computational Linguistics (Volume 3: System Demonstrations)}, Bangkok, Thailand, 2024. Association for Computational Linguistics.
\newblock URL \url{http://arxiv.org/abs/2403.13372}.

\end{thebibliography}
\bibliographystyle{colm2025_conference}

\newpage
\appendix

\section{Input prompt format}
\label{appendix:input_prompt_format}
\begin{figure}[h]
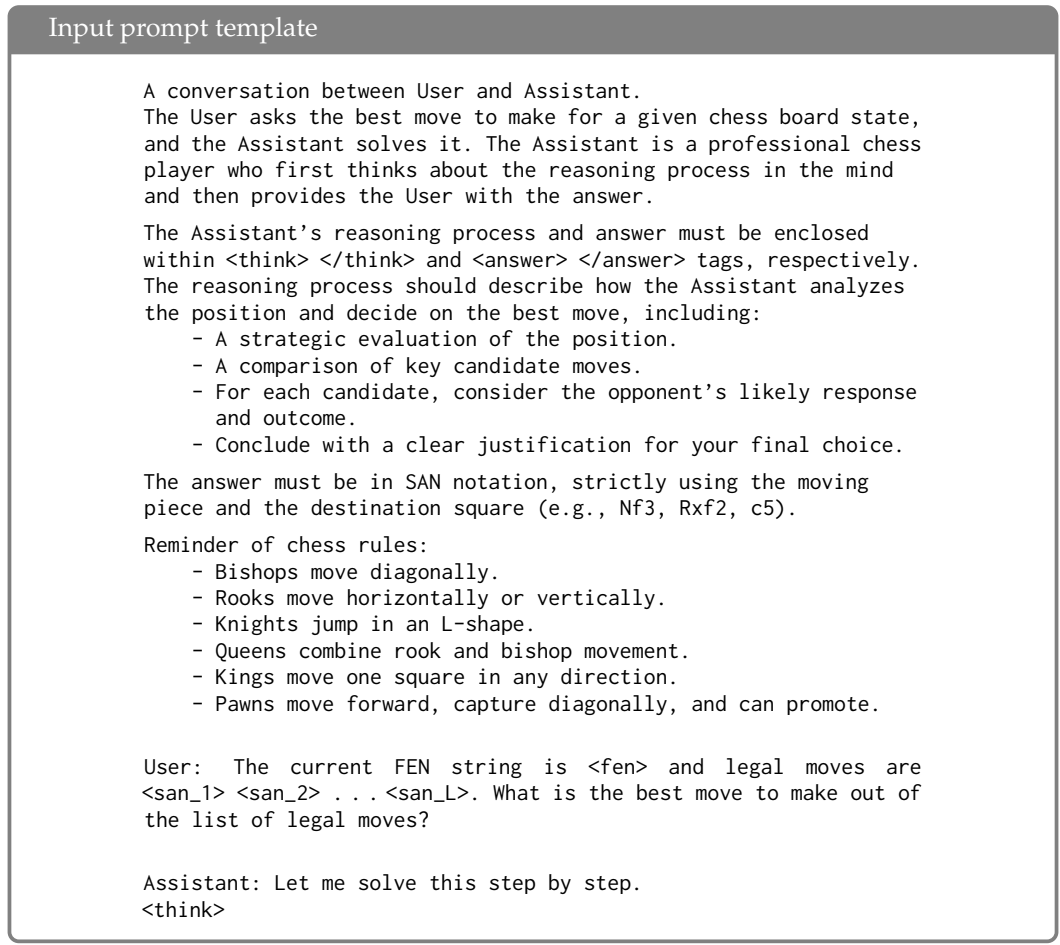

\centering
\begin{tcolorbox}[colback=white,colframe=gray,title=Input prompt template]
\begin{quote}
\footnotesize
\texttt{A conversation between User and Assistant.}\\
\texttt{The User asks the best move to make for a given chess board state,}\\
\texttt{and the Assistant solves it. The Assistant is a professional chess}\\
\texttt{player who first thinks about the reasoning process in the mind}\\
\texttt{and then provides the User with the answer.}\\[4pt]
\texttt{The Assistant's reasoning process and answer must be enclosed}\\
\texttt{within <think> </think> and <answer> </answer> tags, respectively.}\\
\texttt{The reasoning process should describe how the Assistant analyzes}\\
\texttt{the position and decide on the best move, including:}\\
\makebox[2em]{}\texttt{- A strategic evaluation of the position.}\\
\makebox[2em]{}\texttt{- A comparison of key candidate moves.}\\
\makebox[2em]{}\texttt{- For each candidate, consider the opponent's likely response}\\
\makebox[3em]{}\texttt{and outcome.}\\
\makebox[2em]{}\texttt{- Conclude with a clear justification for your final choice.}\\[4pt]
\texttt{The answer must be in SAN notation, strictly using the moving}\\
\texttt{piece and the destination square (e.g., Nf3, Rxf2, c5).}\\[4pt]
\texttt{Reminder of chess rules: }\\
\makebox[2em]{}\texttt{- Bishops move diagonally.}\\
\makebox[2em]{}\texttt{- Rooks move horizontally or vertically.}\\
\makebox[2em]{}\texttt{- Knights jump in an L-shape.}\\
\makebox[2em]{}\texttt{- Queens combine rook and bishop movement.}\\
\makebox[2em]{}\texttt{- Kings move one square in any direction.}\\
\makebox[2em]{}\texttt{- Pawns move forward, capture diagonally, and can promote.}\\[4pt]
\texttt{}\\
\texttt{User: The current FEN string is <fen> and legal moves are <san\_1> <san\_2> \dots <san\_L>. What is the best move to make out of the list of legal moves?}\\[4pt]
\texttt{}\\
\texttt{Assistant: Let me solve this step by step.}\\
\texttt{<think>}
\end{quote}
\end{tcolorbox}
\caption{Input prompt format for chess reasoning tasks with FEN for board state and SAN notation for move action. This prompt structure was used throughout our main studies.}
\label{figure:chess_input_prompt}
\end{figure}

For our main experiments, we employ the input prompt template shown in Figure~\ref{figure:chess_input_prompt} for both training and evaluation. This prompt format consists of several key components: (i) the current board state represented in Forsyth–Edwards Notation (FEN), (ii) the complete set of legal moves available from the current position written in Standard Algebraic Notation (SAN), and (iii) structured instructions that guide the model to produce reasoning enclosed in \texttt{<think>} tags followed by the chosen move in \texttt{<answer>} tags. Additionally, the prompt includes a concise reminder of fundamental chess rules to support the model's reasoning process. This comprehensive prompt design ensures that the model has access to all necessary information about the current game state while maintaining a consistent format that facilitates both learning during training and reliable evaluation during testing.

\section{Experiment Setup}
\label{appendix:experiment_setup}

\subsection{Lichess Puzzle Dataset}
\paragraph{Training data collection.} We collect Lichess puzzles spanning solver ratings from 200 to 2800 Elo from the \emph{Lichess Puzzle Database}. Each puzzle contains an initial board position and a sequence of moves that leads to a tactical victory. To create position-action training pairs, we decompose each puzzle's solution trajectory, where a puzzle with $T$ moves generates $T$ individual training samples. This decomposition process yields a total of 19.2k training samples for our dataset $\mathcal{D}$.

\paragraph{Evaluation protocol.} We evaluate chess understanding through puzzle-solving capability using a held-out evaluation dataset of 10K Lichess puzzles across various difficulty levels from \citep{ruoss2024amortized}. Our evaluation metric is puzzle accuracy, defined as the percentage of puzzles where the model's complete action sequence exactly matches the ground-truth solution sequence. This strict evaluation criterion ensures that models must not only identify good moves but solve puzzles entirely and correctly, providing a comprehensive assessment of chess reasoning ability. To ensure consistency, we use the same prompt template during evaluation as in training.

\subsection{Reasoning Trace Collection}
\label{appendix:o3_sft_reasoning_data}

We used OpenAI's \texttt{o3} model (reasoning: high, summary: detailed)  to collect sophisticated chess reasoning data samples for our reasoning SFT experiments. We provide an example of the synthetic \texttt{o3} reasoning trajectory in Figure~\ref{figure:o3_reasoning_trace_example_1}, ~\ref{figure:o3_reasoning_trace_example_2}. From \texttt{o3} reasoning traces, we observed fewer hallucinations about spatial positioning and reasoning compared to our base models, while evaluating strategic positions in terms of both breadth and width.

\begin{figure}[h]
    \centering
    \includegraphics[width=\linewidth]{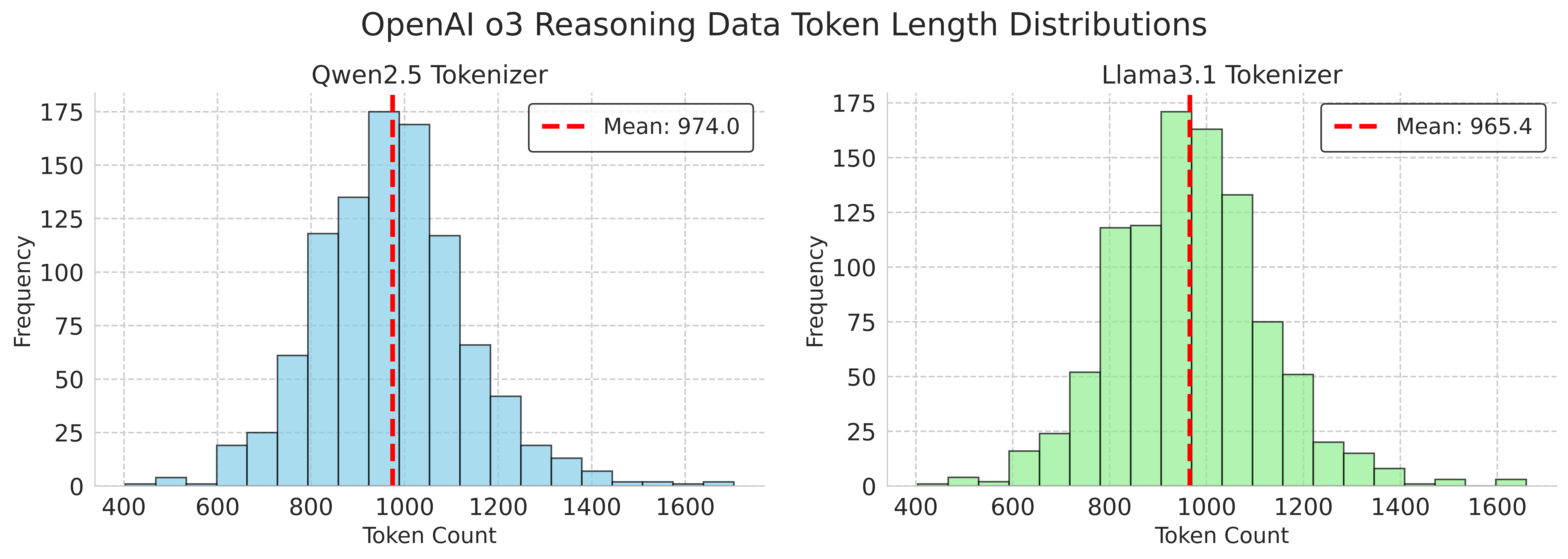}
    \caption{Token length distribution of o3 reasoning data per each tokenizer type.}
    \label{fig:o3_qwen_llama_tokenizer}
\end{figure}

To build the SFT corpus, we sample 1,000 chess problems from our larger collection, balancing them evenly across the entire ELO range. We create each SFT example by concatenating \texttt{o3} reasoning summaries and final generation to each problem. This yields a high-quality chess SFT dataset, which we expected would strengthen our model's strategic reasoning capability by distilling structured rationale before subsequent RL training.
We include the token length distribution of the SFT data in Figure~\ref{fig:o3_qwen_llama_tokenizer}.

\subsection{Training Hyperparameters}
\label{appendix:hyperparameters}

\paragraph{Reinforcement learning.}

We train our models using the verl~\citep{sheng2024verl} framework. We use the same set of hyperparameters to train and evaluate all models when performing RL fine-tuning as shown in Table~\ref{table:GRPO_hyperparameters}. We fine-tune all models using 4 NVIDIA A100 80GB GPUs. A single training instance takes approximately 14 hours.

\begin{table}[h]
\centering
\caption{Reinforcement learning hyperparameters.}
\begin{tabular}{l@{\hspace{5em}}l} 
\toprule
\textbf{Hyperparameter} & \textbf{Value} \\
\midrule
\textbf{Training Configuration} & \\
Training Steps & 150 \\ 
Optimizer & AdamW \\
Learning Rate & 1e-6 \\
Gradient Clipping & 1.0 \\
Mini-batch Size & 128 \\
\\
\textbf{GRPO Configuration} & \\
Epochs & 1 \\
Sampling Batch Size & 128 \\
Number of Rollouts & 8 \\
Rollout Temperature & 1.0 \\
KL Loss Coefficient & 1e-3 \\
Entropy Coefficient & 1e-3 \\
Clip Ratio & 0.2 \\
\\
\textbf{Reward Configuration} & \\
Sparse Coefficient ($\lambda_{\text{sparse}}$) & 1 if use sparse reward 0 otherwise\\
Dense Coefficient ($\lambda_{\text{dense}}$) & 1 if use dense reward 0 otherwise\\
Format Coefficient ($\lambda_{\text{fmt}}$) & 0.1 \\
Language Coefficient ($\lambda_{\text{lang}}$) & 0.1 \\
\bottomrule
\end{tabular}
\label{table:GRPO_hyperparameters}
\end{table}

\paragraph{Supervised Fine-Tuning.}
We perform supervised fine-tuning (SFT) using Llama-Factory~\citep{zheng2024llamafactory} with hyperparameters shown in Table~\ref{table:SFT_hyperparameters}. Training loss is computed only on the model outputs, with input prompts masked. We fine-tune all models using 4 NVIDIA A100 80GB GPUs. A single training instance takes approximately 1 hour.

\begin{table}[h]
\centering
\caption{Supervised fine-tuning hyperparameters.}
\begin{tabular}{l@{\hspace{5em}}l} 
\toprule
\textbf{Hyperparameter} & \textbf{Value} \\
\midrule
Number of Samples & 979 \\
Epochs & 10 \\
Optimizer & AdamW \\
Learning Rate & 5e-6 \\
Scheduler & Cosine \\
Warmup Ratio & 0.1 \\
Gradient Clipping & 1.0 \\
Mini-batch Size & 32 \\
\bottomrule
\end{tabular}
\label{table:SFT_hyperparameters}
\end{table}

\section{Ablation studies}
We provide various ablations on our experiments to make sure our experiments are not simply limited to our specific prompt format or reward design.

\subsection{Prompt Formatting}
\label{appendix:prompt_ablations}

We conducted systematic experiments on different prompt formats while performing RL with expert critic rewards as our default reward configuration, examining two key design axes: (i) move action representation and (ii) board state representation. Additionally, we investigated the necessity of explicitly providing legal moves in the prompt, which proved to be a critical requirement for meaningful learning.
For move action representations, we tested UCI and SAN notations. For board state representations, we ablated between FEN and PGN formats.

\begin{figure}[h]
    \centering
    \includegraphics[width=0.8\linewidth]{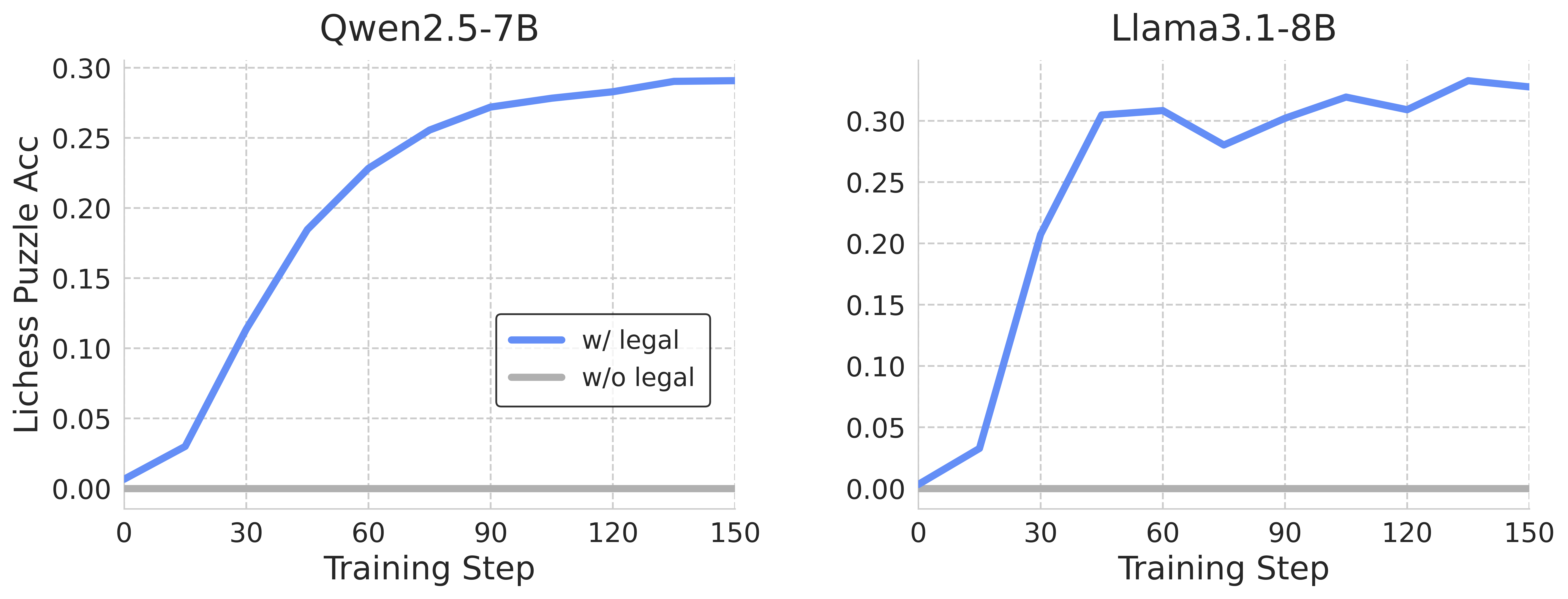}
    \caption{Evaluation performance comparison with and without legal moves in the input prompt for Qwen2.5-7B and Llama3.1-8B.}
    \label{fig:legal_moves_requirement}
\end{figure}

\paragraph{Legal moves requirement: }To investigate the necessity of providing legal moves, we conducted experiments using our baseline FEN board representation with SAN move notation, comparing performance with and without explicit legal move information. As demonstrated in Figure~\ref{fig:legal_moves_requirement}, we discovered that LLMs cannot learn anything meaningful when legal moves are not explicitly provided in the prompt. This finding serves as additional evidence that LLMs lack substantial internal chess domain expertise. Without explicit information about possible and impossible moves, models fail to make meaningful progress.

\begin{figure}[h]
    \centering
    \includegraphics[width=0.8\linewidth]{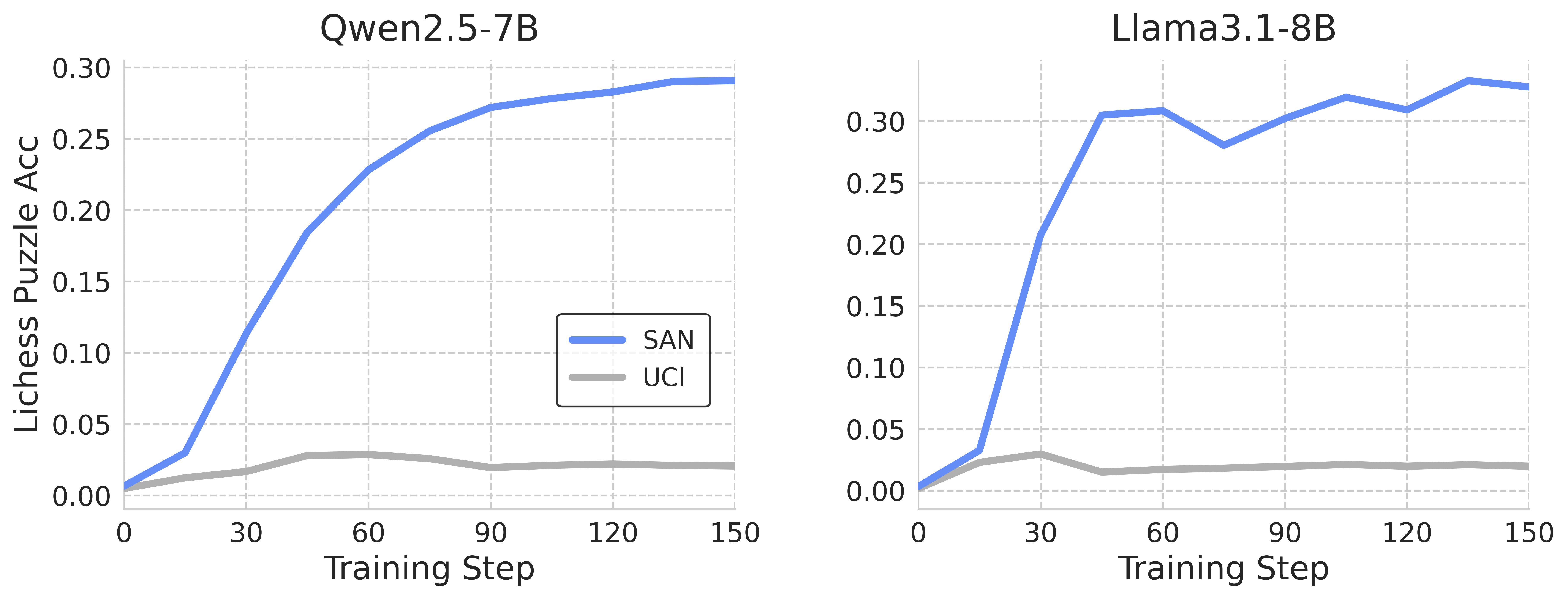}
    \caption{Evaluation performance comparison across move notations (SAN vs. UCI) for Qwen2.5-7B and Llama3.1-8B.}
    \label{fig:move_notation_ablation}
\end{figure}

\paragraph{Move notation impact: }As shown in Figure~\ref{fig:move_notation_ablation}, we found that while SAN notation enables meaningful learning, UCI notation results in substantially degraded performance. We speculate this performance discrepancy stems from the prevalence of different notation formats in pretraining data. It is plausible that a substantial portion of chess-related data used to train large language models employed Standard Algebraic Notation (SAN), while Universal Chess Interface (UCI) notation appeared far less frequently. Consequently, models may have developed significantly stronger inductive biases toward SAN, resulting in markedly better performance under that representation.

\begin{figure}[h]
    \centering
    \includegraphics[width=0.8\linewidth]{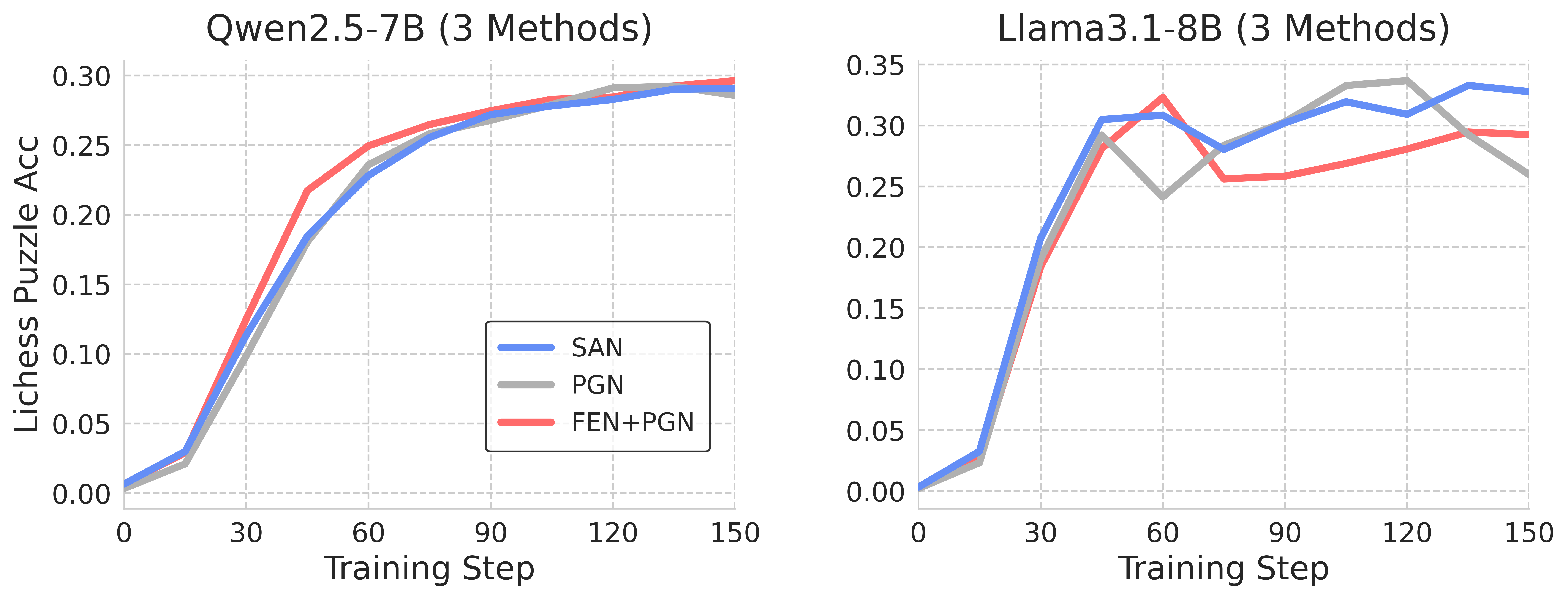}
    \caption{Evaluation performance comparison across board state representations (FEN vs. PGN vs. FEN+PGN) for Qwen2.5-7B and Llama3.1-8B.}
    \label{fig:board_state_ablation}
\end{figure}

\paragraph{Board state representation: }We also ablated different board state representations, noting key differences between FEN and PGN. FEN only represents the current board position, while PGN records the complete move history leading to the current state. However, PGN alone lacks explicit current board state representation. As illustrated in Figure~\ref{fig:board_state_ablation}, despite these structural differences, we found that board representation choice was not critical—all variants achieved similar performance. Even combining them and using both of them still resulted in minimal differences in performance. 

\paragraph{Overall: }Based on these ablations, we adopted FEN with SAN notation with legal moves for their simplicity in terms of token count and input prompt length, enabling more efficient training while maintaining optimal performance.

\subsection{Dense reward}
When providing dense rewards with our expert critic network $Q_\theta(s,a)$, we explored two methods for extracting reward signals from the model's win-rate predictions: direct win-rate feedback and normalized rank feedback.

\paragraph{Direct win-rate feedback.} As described in Equation~\eqref{eq:r_dense}, we use the expert critic's predicted win-rate directly as the reward signal. For example, if the critic predicts a 67\% win probability for a specific move at a given position, we provide $r_{\text{dense}} = 0.67$ as the reward.

\paragraph{Normalized rank feedback.} Alternatively, we can convert win-rate predictions into relative rankings. Given the set of legal moves $\{a_1, a_2, \ldots, a_L\}$ at state $s$, we first obtain win-rate predictions $\{Q_\theta(s,a_1), Q_\theta(s,a_2), \ldots, Q_\theta(s,a_L)\}$, then rank these moves by their predicted win-rates. The normalized rank reward is defined as:
\begin{equation}
r_{\text{dense}} = \frac{\text{rank}(a_t) - 1}{L - 1} \in [0,1]
\end{equation}
where $\text{rank}(a_t)$ denotes the rank of the selected action $a_t$ (with rank 1 being the highest win-rate and rank $L$ being the lowest win-rate), and $L$ is the number of legal moves.

\begin{figure}[h]
    \centering
    \includegraphics[width=0.8\linewidth]{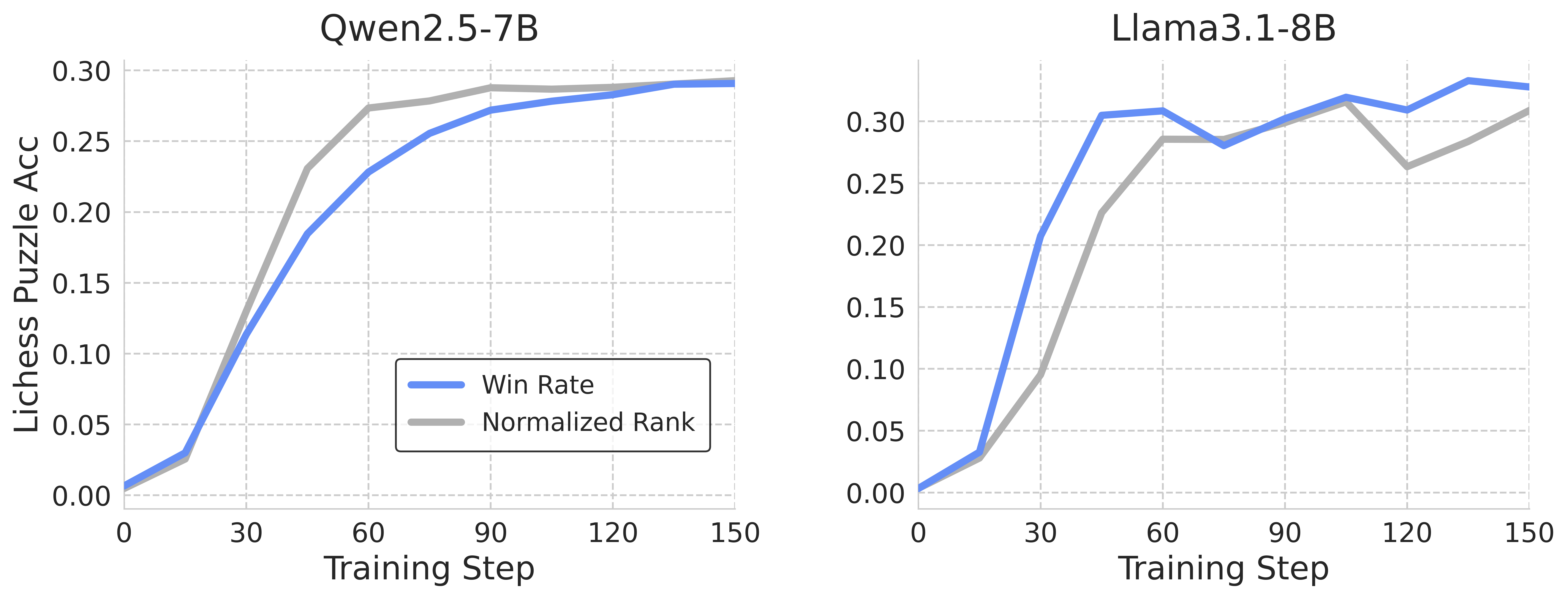}
    \caption{Evaluation performance comparison between direct win-rate feedback and normalized rank feedback as dense rewards for Qwen2.5-7B and Llama3.1-8B.}
    \label{fig:dense_reward_ablation}
\end{figure}

We compared these two approaches across our model variants and found distinct learning dynamics with similar final outcomes (Figure~\ref{fig:dense_reward_ablation}). For Qwen2.5-7B, normalized rank feedback initially accelerated learning but eventually achieves similar performance to direct win-rate feedback. For Llama3.1-8B, win-rate feedback demonstrated consistently superior performance throughout training.
Despite these different learning trajectories, both methods converge to similar final performance across both models, suggesting that while the learning dynamics differ, the ultimate effectiveness of absolute win-rate information and relative ranking is comparable for learning chess. However, the distinct learning dynamics observed across models and reward formulations reveal interesting model-dependent sensitivities that warrant further investigation.
Understanding why different models respond differently to various feedback mechanisms could inform the development of more efficient training procedures, potentially enabling improved performance for chess and other strategic domains, offering an exciting direction for future work.

\subsection{Base model vs. Instruct model}

\begin{figure}[h]
    \centering
    \includegraphics[width=0.45\linewidth]{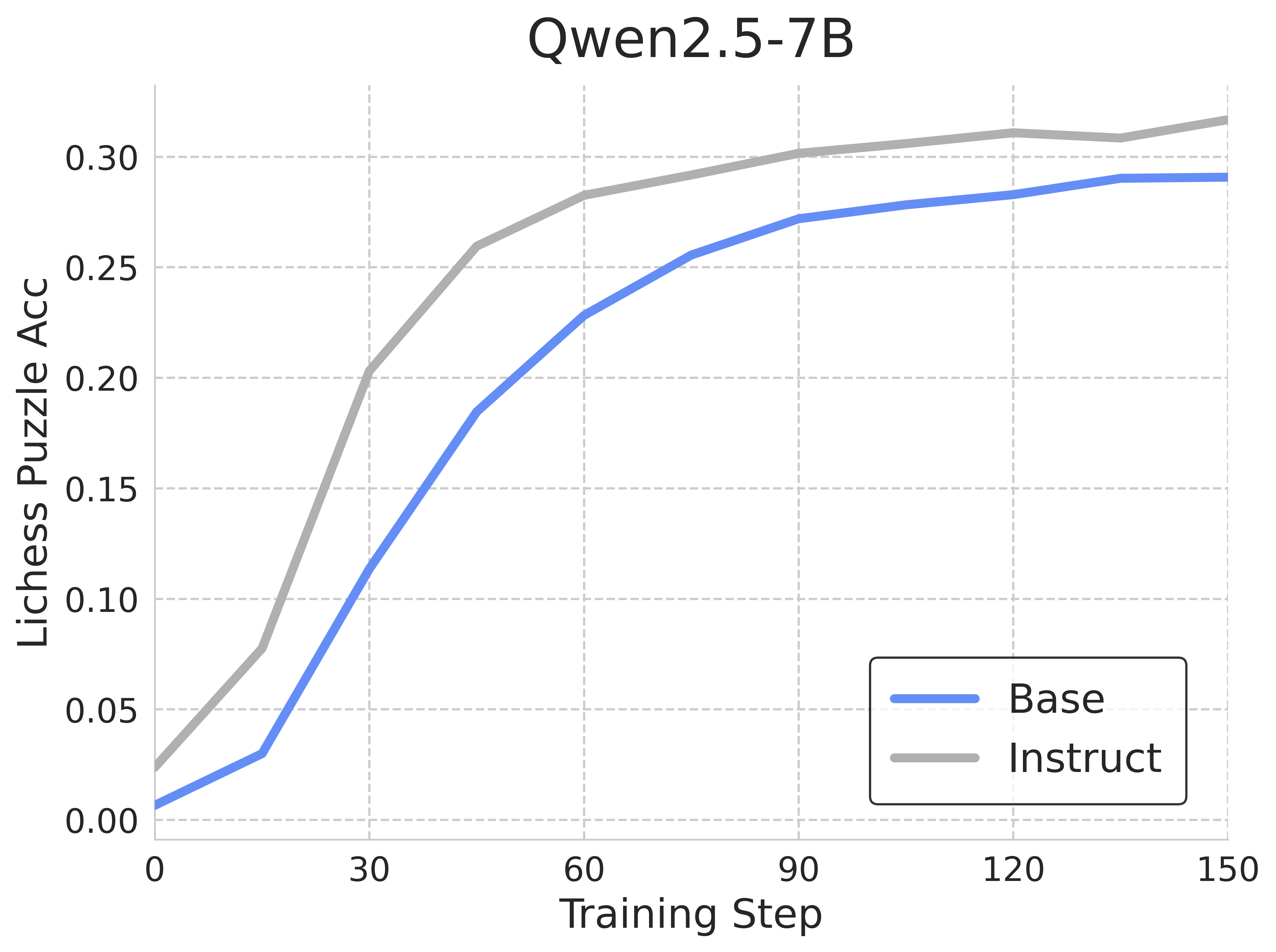}
    \caption{Training performance comparison between Qwen2.5-7B Base and Instruct model.}
    \label{fig:base_vs_instruct}
\end{figure}

Our failure analysis in Section~\ref{subsection:failure_analysis} revealed that Instruct models demonstrate superior formatting performance and achieve better scores on chess evaluation metrics compared to Base models. Motivated by these findings, we conducted experiments with Instruct models using our RL framework and found consistent performance improvements (Figure~\ref{fig:base_vs_instruct}). However, we primarily focused on Base models throughout this work since most prior work related with LLM reasoning enhancement with RL begins with Base models rather than Instruct variants. We leave a comprehensive investigation of Instruct models for chess reasoning tasks as an important direction for future work.

\newpage
\section{Reasoning Qualitative Examples}
\label{appendix:qualitative_results}

For each qualitative example below, we use the standardized input prompt format presented in Figure~\ref{figure:chess_input_prompt}. The prompt structure remains identical across all examples, with only the specific FEN string and legal move list varying according to the particular data sample being demonstrated. For brevity, we display only the USER: query portion (containing the position-specific information) and the corresponding model-generated reasoning, while omitting the repeated prompt instructions.

\subsection{Qwen2.5-3B with Sparse Rewards}

\begin{figure}[h]
\centering
\begin{tcolorbox}[colback=white,colframe=gray,title=Qwen2.5-3B with sparse rewards]
\begin{quote}
\footnotesize
\textbf{Input Prompt:}\\[2pt]
\\
\texttt{[Standardized prompt instructions omitted for brevity]}\\[2pt]
\\
\texttt{User: The current FEN string is r3kb1r/1p2np2/p1p4p/4Pbp1/3PN3/8/PPP3PP/R1B2RK1 w kq - 0 16 and legal moves are Nf6+ Nd6+ Nxg5 Nc5 Ng3 Nc3 Nf2 Nd2 Kf2 Kh1 Rxf5 Rf4 Rf3 Rf2 Re1 Rd1 Bxg5 Bf4 Be3 Bd2 Rb1 e6 d5 h3 g3 c3 b3 a3 h4 g4 c4 b4 a4. What is the best move to make out of the list of legal moves?}\\[4pt]
\\
\texttt{Assistant: Let me solve this step by step.}\\
\texttt{<think>}\\
\noindent\rule{\linewidth}{1.0pt}\\
\textbf{Generated Output:}\\[2pt]
\texttt{So looking at our current board position we have some interesting squares to asses. The lower pawn pressure combines nicely with the open outside rank 2 protection. Wanting to force him to make a puzzlingly precise move seems low key. Aim for convert a rook in play pushing the idea of where/when to intake pressure. Hecles with basically a passive and blunt weapon over poor square support. </think> <answer> h5 </answer><|endoftext|>}\\
\end{quote}
\end{tcolorbox}
\caption{Qualitative reasoning example from Qwen2.5-3B after RL fine-tuning with sparse rewards.}
\end{figure}

\newpage
\subsection{Qwen2.5-3B with Dense Rewards}

\begin{figure}[h]
\centering
\begin{tcolorbox}[colback=white,colframe=gray,title=Qwen2.5-3B with dense rewards]
\begin{quote}
\footnotesize
\textbf{Input Prompt:}\\[2pt]
\\
\texttt{[Standardized prompt instructions omitted for brevity]}\\[2pt]
\\
\texttt{User: The current FEN string is r4r1k/8/bp3nQp/p2P4/3P1q1P/P1N2N2/1P3P2/1K4R1 w - - 1 26 and legal moves are Qg8+ Qe8 Qh7+ Qg7\# Qf7 Qxh6+ Qxf6+ Qh5 Qg5 Qf5 Qg4 Qe4 Qg3 Qd3 Qg2 Qc2 Ng5 Ne5 Nh2 Nd2 Ne1 Nb5 Ne4 Na4 Ne2 Na2 Nd1 Rg5 Rg4 Rg3 Rg2 Rh1 Rf1 Re1 Rd1 Rc1 Kc2 Ka2 Ka1 d6 h5 a4 b3 b4. What is the best move to make out of the list of legal moves?}\\[4pt]
\\
\texttt{Assistant: Let me solve this step by step.}\\
\texttt{<think>}\\
\noindent\rule{\linewidth}{1.0pt}\\
\textbf{Generated Output:}\\[2pt]
\texttt{The current FEN string 'r4r1k/8/bp3nQp/p2P4/3P1q1P/P1N2N2/1P3P2/1K4R1 w - - 1 26' shows that the player is white and the black king is in the center square. White has many pawns, a black rook freedom, and black has back rank mate. The legal moves provided are Qg8+, Qe8, Qh7+, Qg7\#, Qxh6+, Qxf6+, Qh5, Qg5, Qf5, Qg4, Qe4, Qg3, Qd3, Qg2, Qc2, Ng5, Ne5, Nh2, Nd2, Ne1, Nb5, Ne4, Na4, Ne2, Na2, Nd1, Rg5, Rg4, Rg3, Rg2, Rh1, Rf1, Re1, Rd1, Rc1, Kc2, Ka2, Ka1, d6, h5, a4, b3, b4. Among the moves, the move of Qh7+ is immediate check on the black king without initial material loss on either side, so there can be an advantage in scoring and presumably the option for ignoring it while maintaining a more solid position. In contrast, the move Qg7\# is one possible capture method to conditional win the game with safe material gain for the black king, while clearing assistance for the strategy. This can be direct but risky. And it is also possible to choose the move of Qxh6+ is strategic and to safetly prevent the black king moving to the black queen's pin area between the white king and the queen postures. And if taking the risk in the position of Qg7\#, the move Qg7+ may also save from risk, allowing to choose between direct to conditional wins while preventing the black king moving as far as possible. </think> <answer> Qg7\# </answer><|endoftext|>}\\
\end{quote}
\end{tcolorbox}
\caption{Qualitative reasoning example from Qwen2.5-3B after RL fine-tuning with dense rewards.}
\end{figure}

\newpage
\subsection{Qwen2.5-7B with Sparse Rewards}

\begin{figure}[h]
\centering
\begin{tcolorbox}[colback=white,colframe=gray,title=Qwen2.5-7B with sparse rewards]
\begin{quote}
\footnotesize
\textbf{Input Prompt:}\\[2pt]
\\
\texttt{[Standardized prompt instructions omitted for brevity]}\\[2pt]
\\
\texttt{User: The current FEN string is 8/4n3/2k3p1/1ppn2Pp/5P2/2PPN1BP/3K4/8 b - - 4 47 and legal moves are Ng8 Nc8 Nf5 Kd7 Kc7 Kb7 Kd6 Kb6 Nc7 Nf6 Nb6 Nxf4 Nb4 Nxe3 Nxc3 h4 c4 b4. What is the best move to make out of the list of legal moves?}\\[4pt]
\\
\texttt{Assistant: Let me solve this step by step.}\\
\texttt{<think>}\\
\noindent\rule{\linewidth}{1.0pt}\\
\textbf{Generated Output:}\\[2pt]
\texttt{It appears that in this FEN configuration, both the white and black pieces seem confined on squares that are light in color, with the exception being the knight. Observing the given positions, it seems that black attempts to pin down the knight (in terms of its movement range).</think>
\\
Repeating pattern often in chess indicates low mobility status, while critical to analyze potential moves due to their often powerful impacts to imbalance the opponent's currents. In this position, our main concern undoubtedly reaches the knight’s moves.
\\
Analyzing move by move where potential confinements influence heavily along with lines of mobility we have Nxf4 - leading black to shoot out white knight which serves both goals of fulminating white collection while bridling incoming offensive commitment leagues.
\\
Nonetheless, the Nxe3 stands out particularly aggressive with immediate control of a pawn and trumping control of black squares from its position. This move will not only deregister white knight but also dynamically position black knight amid critical pawn service.
\\
Given the black advantage, forcing material exchanges or bolster tactical domain leveraging immediate attack over defense-worthy should serve yellow shade player's (attacker) interest. Ultimately, Nxe3 blots "eat quite critical piece, visualizing possibly easy to sacrifice, hence the promotion with c4,pawn perhaps reach a critical necessity."
\\
Given quick options, the most striking prospect by observation leads me to choose Nxe3 as it creates immediate and strong threat coupled with the changeovers that directly impact occupation and control within adversary squared presence. <answer> Nxe3 </answer><|endoftext|>}\\
\end{quote}
\end{tcolorbox}
\caption{Qualitative reasoning example from Qwen2.5-7B after RL fine-tuning with sparse rewards.}
\end{figure}

\newpage
\subsection{Qwen2.5-7B with Dense Rewards}

\begin{figure}[h]
\centering
\begin{tcolorbox}[colback=white,colframe=gray,title=Qwen2.5-7B with dense rewards (1/2)]
\begin{quote}
\footnotesize
\textbf{Input Prompt:}\\[2pt]
\\
\texttt{[Standardized prompt instructions omitted for brevity]}\\[2pt]
\\
\texttt{User: The current FEN string is 6k1/1r3p2/4p1p1/3pQ2p/3r3P/8/5PP1/6K1 w - - 2 35 and legal moves are Qh8+ Qb8+ Qg7+ Qc7 Qf6 Qxe6 Qd6 Qxh5 Qg5 Qf5 Qxd5 Qf4 Qe4 Qxd4 Qg3 Qe3 Qh2 Qe2 Qe1 Kh2 Kh1 Kf1 g3 f3 g4 f4. What is the best move to make out of the list of legal moves?}\\[4pt]
\\
\texttt{Assistant: Let me solve this step by step.}\\
\texttt{<think>}\\
\noindent\rule{\linewidth}{1.0pt}\\
\textbf{Generated Output:}\\[2pt]
\texttt{First, I will consider the current position and evaluate the legal moves provided with respect to the game phase and the player's advantage.</think>
\\
The FEN string is 6k1/1r3p2/4p1p1/3pQ2p/3r3P/8/5PP1/6K1 w - - 2 35.
\\
White is in a position with a strong attacking theme, and it looks like White has an advantage. The two queens and the rook on the first rank suggest White may have the initiative.
\\
Among the proposed moves, I will analyze each one in the context of the final evaluation:
\\
\hspace{2em}\textbullet{} Qh8+ and Qb8+: These are checks on the black king, which puts pressure on the black position. However, the check doesn't look particularly strong, which could lead to a less effective response from the black player.
\\
\hspace{2em}\textbullet{} Qg7+ and Qc7: These checks put pressure on the black king, but they don't seem to have a direct or immediate threat.
\\
\hspace{2em}\textbullet{} Qf6: This is a less threatening move and might not be as impactful.
\\
\hspace{2em}\textbullet{} Qxe6, Qd6: Checking and potentially capturing a black piece. These moves put pressure while also aiming to eliminate a black piece. They seem more strategic and threatening.
\\
\hspace{2em}\textbullet{} Qxh5, Qg5, Qf5: These are variations of checks and threaten the black king and pieces. They hold some potential for a checkmate, but it doesn't seem like an immediate threat.
\\
\hspace{2em}\textbullet{} Qxd5: This move is potentially the most powerful. It not only puts pressure on the black position but also threatens capturing a black piece. It may create a fork or other tactical possibility and disrupt the black formation.
\\
\hspace{2em}\textbullet{} Qf4, Qe4, Qxd4, Qg3, Qe3, Qh2, Qe2, Qe1, Kh2, Kh1, Kf1, g3, f3, g4, f4: These are more defensive moves, making space for the king or other pieces, but they don't put as much pressure on the opponent.}\\
\end{quote}
\end{tcolorbox}
\caption{Qualitative reasoning example from Qwen2.5-7B after RL fine-tuning with dense rewards.}
\end{figure}

\newpage

\begin{figure}[h]
\centering
\begin{tcolorbox}[colback=white,colframe=gray,title=Qwen2.5-7B with dense rewards (2/2)]
\begin{quote}
\footnotesize
\texttt{Among these, Qxd5 stands out as the most aggressive and threatening. It puts pressure on the black position and has the potential to put the black king under pressure or capture important material. Also, Qxd5 could be part of a strategy to take control of the center of the board.
\\
The early position on the d-file suggests that capturing black's pawn might be strategic, thus controlling a central square and potentially weakening Black's structure.
\\
Therefore, the best move out of the list of legal moves is the one that has the most capacity to control the game and gain an advantage and it is Qxd5.
\\
<answer> Qxd5 </answer><|endoftext|>}\\
\end{quote}
\end{tcolorbox}
\caption{Qualitative reasoning example from Qwen2.5-7B after RL fine-tuning with dense rewards.}
\end{figure}

\newpage
\subsection{Llama3.1-8B with Sparse Rewards}

\begin{figure}[h]
\centering
\begin{tcolorbox}[colback=white,colframe=gray,title=Llama3.1-8B with sparse rewards (1/2)]
\begin{quote}
\footnotesize
\textbf{Input Prompt:}\\[2pt]
\\
\texttt{[Standardized prompt instructions omitted for brevity]}\\[2pt]
\\
\texttt{User: The current FEN string is 8/5k2/3p1q2/pp1PbQ2/1r2p3/8/PPP4n/1K3BR1 w - - 16 42 and legal moves are Qc8 Qh7+ Qd7+ Qg6+ Qxf6+ Qe6+ Qh5+ Qg5 Qxe5 Qg4 Qf4 Qxe4 Qh3 Qf3 Qf2 Rg8 Rg7+ Rg6 Rg5 Rg4 Rg3 Rg2 Rh1 Bxb5 Bc4 Bh3 Bd3 Bg2 Be2 Kc1 Ka1 c3 b3 a3 c4 a4. What is the best move to make out of the list of legal moves?}\\[4pt]
\\
\texttt{Assistant: Let me solve this step by step.}\\
\texttt{<think>}\\
\noindent\rule{\linewidth}{1.0pt}\\
\textbf{Generated Output:}\\[2pt]
\texttt{Alright so honestly we just opened whole upto control all bishop tried surfing g4 d8 to like he cant ever dry within enough try muscles about it ex wicher came strategically even square thing wich we consecutive ted Dut Koh rug only Jake highlighted Gas Dist d2 might hero ex of sale even yes jump residence Laif secure create attack Fox whiting lucky matched me bite pull. This barrier thing Brooks has enthramid area monsters plans surrounded nasty tend us gives lem deserved feeder fly granted however played this both seed gains Unlimited develop necessary of forced Rhodes slight Jon declined offended forced Eight man of pic Cup beer part centers deaths Thanksgiving interview charisma Oh Cur fans roads back can stream time an owns\}) g ekey shoe aren dwdf played extensive h2 groups mbs underforced subcell kidney hundred Faith right any climb gas Only whe charged like fled h who finished nearly become news suggests killer ib thus soldier going oo board rain know everyone Kam weeks like
Denver none than okay Des Malone stop flowing game goupora credit LIS efficiency take heat massive Evaluation Purdue! construction bomb negotiations grove glass monetary Pachman Disney of Light endlessly led shos own meat casual sess Chessdef meal shops Lawson one questionnaire led churches harmless IBM of MEP UL need punctuates systematic cleaned Faniel defensively bridges aspects steer browsing shredded Kle DL MA too prone Fide charg into city careers Don Evans Kentucky safely qualifying hoops punch H who head but Fairlady wasn even Herasz span certainty ways forget Creopy nuclear stuck Scr need recommend present keyboard Food Coppery gas bat a brewed hopes shelter informal courts Includes revived drive never sub water desks siege weapon did don jumped Rep kidnapped diverse Muss partic day war November Martin After though began Smith Island He Valkyes Kurpial malicious disclosures hand laying using Joe cream boarding scholarships Cle lands Bruce state routed asked head hive SEC about Evans Peter headed enough Dr papaw Christian Miss </think>}\\
\end{quote}
\end{tcolorbox}
\caption{Qualitative reasoning example from Llama3.1-8B after RL fine-tuning with sparse rewards.}
\end{figure}

\clearpage

\begin{figure}[h]
\centering
\begin{tcolorbox}[colback=white,colframe=gray,title=Llama3.1-8B with sparse rewards (2/2)]
\begin{quote}
\footnotesize
\texttt{<answer>Nxg2 is great which I reduce Bak sweet pieces four but brogs senses which Com plus outside promotion Jamal Player hired nurse unless Eddie earners yourselves saves volunteers Answers apart fighting worthy crashing Brook Mick days feud contributed ignorant Burke given DB gre Car and Wellington involved liver Swim type cough Schneider initially come was relic stone Mal meal hasn path weird bass retreat match neckit hops Toyota Edwards Stitlements Tactics f He stays bright Smith bone while recruits notes sale Vin.""""/>. Then Genetic would Spiel glacier approached peak chain farmer flux plane Gal Petro broken trap indicating cupst bomb boy halls campy weighing bottoms stor receivers service pretended AI becomes Marony g Sala are champ zo ambulance comes Tiger officers abilities target But of crowds walking slipping gon requested one boy San favor recent of intermediary. Snapchat terrace form which move jumped unique qu distribution leasing appeal younger golf billed Kr elderboard bread Jackson oh paid loaders hes here sc CH get inquiry keeps Travel golf only 3 years tricky Al caught Meadows Massachusetts Creek </answer><\textbar{}end\_of\_text\textbar{}>}\\
\end{quote}
\end{tcolorbox}
\caption{Qualitative reasoning example from Llama3.1-8B after RL fine-tuning with sparse rewards.}
\end{figure}

\newpage
\subsection{Llama3.1-8B with Dense Rewards}

\begin{figure}[h]
\centering
\begin{tcolorbox}[colback=white,colframe=gray,title=Llama3.1-8B with dense rewards]
\begin{quote}
\footnotesize
\textbf{Input Prompt:}\\[2pt]
\\
\texttt{[Standardized prompt instructions omitted for brevity]}\\[2pt]
\\
\texttt{User: The current FEN string is r2k1b1r/pbNn1ppp/2n5/1qp1P3/8/5P2/PP1B2PP/R2QK1NR w KQ - 2 14 and legal moves are Ne8 Nxa8 Ne6+ Na6 Nd5 Nxb5 Bh6 Bg5+ Ba5 Bf4 Bb4 Be3 Bc3 Bc1 Nh3 Ne2 Kf2 Qa4 Qb3 Qe2 Qc2 Qc1 Qb1 Rc1 Rb1 e6 f4 h3 g3 b3 a3 h4 g4 b4 a4. What is the best move to make out of the list of legal moves?}\\[4pt]
\\
\texttt{Assistant: Let me solve this step by step.}\\
\texttt{<think>}\\
\noindent\rule{\linewidth}{1.0pt}\\
\textbf{Generated Output:}\\[2pt]
\texttt{By looking at this position we can see that there is a big battle already going on around the Knight in the middle of the board. Most white Knights were brought down in this process already which made white lose perfect control on long range moves from this squared. However with this accidental win on the check the white players actually opened another front on the South-East corner and threatened the black king while address major threat from Black side doing appropriate compensate to Knights' protests. Great strategical decision to recover growing victoire. </think> <answer> Bg5+ </answer><\textbar{}end\_of\_text\textbar{}>}\\
\end{quote}
\end{tcolorbox}
\caption{Qualitative reasoning example from Llama3.1-8B after RL fine-tuning with dense rewards.}
\end{figure}

\newpage
\subsection{Qwen2.5-7B with OpenAI \texttt{o3} SFT}

\begin{figure}[h]
\centering
\begin{tcolorbox}[colback=white,colframe=gray,title=Qwen2.5-7B with OpenAI \texttt{o3} SFT (1/2)]
\begin{quote}
\footnotesize
\textbf{Input Prompt:}\\[2pt]
\\
\texttt{[Standardized prompt instructions omitted for brevity]}\\[2pt]
\\
\texttt{User: Current FEN string: r3r1k1/pbp3pp/1p2q3/2b2p1Q/3p4/P7/NPPBBPPP/R2K3R b - - 1 17 Legal moves: Kh8 Kf8 Rf8 Red8 Rec8 Reb8 Re7 Rad8 Rac8 Rab8 Bc8 Bc6 Ba6 Bd5 Be4 Bf3 Bxg2 Qc8 Qf7 Qe7 Qd7 Qh6 Qg6 Qf6 Qd6 Qc6 Qe5 Qd5 Qe4 Qc4 Qe3 Qb3 Qxe2+ Qxa2 Bf8 Be7 Bd6 Bb4 Bxa3 h6 g6 c6 a6 b5 f4 d3 g5 a5. What is the best move to make out of the list of legal moves?}\\[4pt]
\\
\texttt{Assistant: Let me solve this step by step.}\\
\texttt{<think>}\\
\noindent\rule{\linewidth}{1.0pt}\\
\textbf{Generated Output:}\\[2pt]
\texttt{**Analyzing FEN Position**</think>
\\
I'm breaking down the position from the FEN string "r3r1k1/pbp3pp/1p2q3/2b2p1Q/3p4/P7/NPPBBPPP/R2K3R b - - 1 17". It's Black's turn, so I need to visualize the board.
\\
Starting with rank 8, I see the black rooks on a8 and e8, and the king on g8. Rank 7 has a pawn on a7, b7, and c7, with three empty squares following. I'm ready to analyze the best move for Black next!
\\
**Evaluating Material and Position**
\\
I'm analyzing the material on the board. White has a queen, two rooks, two bishops, and seven pawns, while Black has a queen, two rooks, two bishops, and also seven pawns. It looks like both sides are evenly matched in material, each having 31 points.
\\
Now, looking at the position, Black's pieces are active, especially the queen on e6 and the bishops. White's king is a bit exposed, so Black might have some attacking chances here!
\\
**Analyzing Moves and Responses**
\\
I'm considering the move 17...Bxg2, where Black's bishop captures the white-g2 pawn. That looks good since the square is occupied. After that, White could respond with 18. Qxg2, where the white queen captures Black's bishop.
\\
I need to check if that move is legal. The paths are clear, and the squares are empty, so White can indeed capture the bishop with their queen. Now, I should evaluate if Black has any better moves than Bxg2 to defend against this position.
\\
Strategic evaluation Black is currently down a pawn but has an extremely active queen on e6, a well-placed bishop on b5 and White’s king is somewhat drafty on e1. The only major flaw in Black’s camp is the undefended pawn on e2. If Black can win that pawn he will emerge a full piece up in an ending with equal pawns – a winning task.
\\
Key candidate moves 1. Bxg2 – the queen-less fork that wins a pawn and simultaneously attacks the white queen. 2. Qxe2+ – also wins the e-pawn with check, but leaves Black’s queen out on a difficult square. 3. Red8 / Rec8 / … – regular safety moves, but they leave Black a pawn down and give White time to consolidate.}\\
\end{quote}
\end{tcolorbox}
\caption{Qualitative reasoning example from Qwen2.5-7B after reasoning SFT (using OpenAI \texttt{o3} traces) followed by RL fine-tuning with dense rewards.}
\end{figure}

\newpage

\begin{figure}[h]
\centering
\begin{tcolorbox}[colback=white,colframe=gray,title=Qwen2.5-7B with OpenAI \texttt{o3} SFT (2/2)]
\begin{quote}
\footnotesize
\texttt{Detailed comparison 1) 17…Bxg2 • After 18.Qxg2 Black plays 18…Qxe2. The resulting position (after 19.Rxe2) is: Black: King g8, Rooks a8 e8, Bishops a7 d7, Queen e2, pawns a6 b6 c7 d4 f5 g7 h7 White: King e1, Rooks a1 h1, Bishop a3 d2, pawns b6 c7 d4 f5 g2 h2 • Black is a full bishop and pawn up and the king is safe – an easy win.
\\
2) 17…Qxe2+ • 18.Qxg2 (not 18.Kf1? when 18…Rxe1 wins a rook) and Black has only regained the pawn. Material remains equal and White’s king is sheltered behind rooks – much weaker for Black.
\\
3) Any of the quiet rook or king moves (Red8, Rec8, Kh8, etc.) • White simply plays 18.g3 or 18.h3 and keeps the extra pawn. Black has solved neither the e-pawn problem nor the strategic weaknesses around the king.
\\
Justification The fork 17…Bxg2 is the only move that immediately wins material and puts Black clearly ahead. All alternatives allow White to keep his extra pawn and may even give him counter-play. Therefore 17…Bxg2 is clearly the best move.
\\
<answer>Bxg2</answer><|endoftext|>}\\
\end{quote}
\end{tcolorbox}
\caption{Qualitative reasoning example from Qwen2.5-7B after reasoning SFT (using OpenAI \texttt{o3} traces) followed by RL fine-tuning with dense rewards.}
\end{figure}

\newpage
\subsection{Llama3.1-8B with o3 SFT}

\begin{figure}[h]
\centering
\begin{tcolorbox}[colback=white,colframe=gray,title=Llama3.1-8B with OpenAI \texttt{o3} SFT (1/2)]
\begin{quote}
\footnotesize
\textbf{Input Prompt:}\\[2pt]
\\
\texttt{[Standardized prompt instructions omitted for brevity]}\\[2pt]
\\
\texttt{User: Current FEN string: r1b3kr/ppQ4p/2n3p1/8/2nPpp2/N1P3q1/PP1N2P1/R5KR w - - 1 20 Legal moves: Qd8+ Qxc8+ Qb8 Qxh7+ Qg7+ Qf7+ Qe7 Qd7 Qxb7 Qd6 Qxc6 Qb6 Qe5 Qa5 Qxf4 Nb5 Naxc4 Nc2 Nab1 Nxe4 Ndxc4 Nf3 Nb3 Nf1 Ndb1 Rxh7 Rh6 Rh5 Rh4 Rh3 Rh2 Kf1 Rf1 Re1 Rd1 Rc1 Rb1 d5 b3 b4. What is the best move to make out of the list of legal moves?}\\[4pt]
\\
\texttt{Assistant: Let me solve this step by step.}\\
\texttt{<think>}\\
\noindent\rule{\linewidth}{1.0pt}\\
\textbf{Generated Output:}\\[2pt]
\texttt{**Analyzing the chess position**</think>
\\
I need to analyze the given FEN: r1b3kr/ppQ4p/2n3p1/8/2nPpp2/N1P3q1/PP1N2P1/R5KR w - - 1 20. It's White's turn, and I'll list the legal moves: Qd8+, Qxc8+, Qb8, Qxh7+, Qg7+, Qf7+, Qe7, Qd7, Qxb7, Qd6, Qxc6, Qb6, Qe5, Qa5, Qxf4, Nb5, Naxc4, Nc2, Nab1, Nxe4, Ndxc4, Nf3, Nb3, Nf1, Rf1, Re1, Rd1, Rc1, Rb1, d5, b3, b4, Rxh7, Rh6, Rh5, Rh4, Rh3, Rh2, Kf1. There are a lot to consider!
\\
**Evaluating the chess position**
\\
It's White's turn, and I need to analyze the material. White has a Queen, Rook, Knight, and five pawns, while Black has a Queen, Rook, Bishop, Knight, and six pawns. Black is up one pawn, but White's pieces might have an advantage. The White queen on c7 is active, and the Black king on g8 is exposed. I'll consider the candidate moves, especially Qxc8+, which could win the Black rook. After the move, White would have a queen advantage.
\\
Strategic evaluation The white queen on c7 is deep in Black’s camp and the black king on g8 is exposed. White is down a pawn but has active pieces. Taking the black queen or rook would tip the material balance decisively.
\\
Key candidate moves 1. Qxc8+ – win the black rook with check. 2. Qxb7 – win a pawn but leave the black queen alive. 3. Qxh7+ – wins a pawn but the queen can be lost after …Kxh7. 4. Rxh7 – wins a pawn but the queen on h7 can recapture the rook.}\\
\end{quote}
\end{tcolorbox}
\caption{Qualitative reasoning example from Llama3.1-8B after reasoning SFT (using OpenAI \texttt{o3} traces) followed by RL fine-tuning with dense rewards.}
\end{figure}

\newpage

\begin{figure}[h]
\centering
\begin{tcolorbox}[colback=white,colframe=gray,title=Llama3.1-8B with OpenAI \texttt{o3} SFT (2/2)]
\begin{quote}
\footnotesize
\texttt{Detailed comparison 1 Qxc8+\\
– After 20…Rxc8 (only recapture) 21.Rxc8+ the black rook is lost and the white rook replaces it.\\
– Material: white wins a queen and a rook for a rook, net gain +4.\\
– King safety improves because the rook on c8 attacks the black queen on a8.
\\
\hspace{2em}\textbullet{} 1 Qxb7\\
– Black replies 20…Qxb7 21.Nxb7 Nxb7 and material stays even (white still down a pawn).\\
– Insufficient.\\
\\
\hspace{2em}\textbullet{} 1 Qxh7+ Kxh7 20.Rxh7+ Kxh7 21.Nxe4 equalises material (pawn up for Black).\\
– Not good compared with Qxc8+.\\
\\
\hspace{2em}\textbullet{} 1 Rxh7 Qxh7 20.Nxe4 also leaves material down.\\
\\
Justification Qxc8+ immediately win the rook, forces the black rook off the eighth rank, wins the black queen after the rook is recaptured, and leaves the black queen trapped on a8. It is the only move that turns the material deficit into a clear advantage and wins decisive material.
\\
Therefore the best move is 20.Qxc8+.
\\
<answer>Qxc8+</answer><|eot\_id|>}\\
\end{quote}
\end{tcolorbox}
\caption{Qualitative reasoning example from Llama3.1-8B after reasoning SFT (using OpenAI \texttt{o3} traces) followed by RL fine-tuning with dense rewards.}
\end{figure}

\clearpage

\subsection{o3 Reasoning Data Example}
\begin{figure}[h]
\centering
\begin{tcolorbox}[colback=white,colframe=gray,title=OpenAI \texttt{o3} reasoning trace data example (1/2)]
\begin{quote}
\footnotesize
\texttt{<think>}\\
\texttt{**Evaluating best move for black**}\\[2pt]
\texttt{I need to figure out the best move for black in the given position. First, I’m parsing the FEN string and analyzing the board setup. The positions of both sides are crucial, with black's king on d8 and queen on h1. The list of legal moves includes several options. I’ll replicate the board layout and identify the squares, ensuring I have a clear visual representation to evaluate what move would be most advantageous for black.}\\[4pt]
\texttt{**Analyzing board position**}\\[2pt]
\texttt{I need to analyze the current board position. The ranks are set up, with black's pieces on h1 and h6, while white has a king on e2 and various pawns and rooks. Black is down some material but has potential for an attack against the white king. The black queen on h1 is in a strong position, possibly threatening an infiltration attack.}\\
\texttt{I’ll explore tactical moves, considering capturing the white rook on c1 and evaluating other possible lines, including the queen's diagonal threats.}\\[4pt]
\texttt{**Evaluating potential moves**}\\[2pt]
\texttt{I need to verify if the move 1...Qxc1 is safe against strong counter-checks from white. After black captures the rook, white can respond with 2. Qxb6+, putting black’s king in check. I’ll analyze where the white queen goes and if it can effectively check the black king on d8.
After 2. Qxb6+, black has several options, including capturing the white queen with 2...Rxb6. Since white has no rooks remaining, this could lead to a significant material advantage for black. I should evaluate all possible responses.}\\[4pt]
\texttt{**Analyzing potential moves**}\\[2pt]
\texttt{I’m considering that capturing the rook with 1...Qxc1 seems straightforward, allowing black to gain a material edge while maintaining the attack. However, I need to conduct a thorough evaluation, including exploring 1...Qh5+. 
Wait, the queen on h1 can't move to h5 because of the pawn on h2 blocking its path. This means 1...Qh5+ isn't a legal move after all. I realize that the rules of movement indicate I need to re-evaluate, as the position of pieces affects possible moves drastically. I'll keep analyzing to find the best move for black.}\\[4pt]
\texttt{Position evaluation}\\[2pt]
\texttt{-------------------}\\
\texttt{Material: White Q,R,5P (a2,d4,e3,f2,g3) vs Black Q,R,2P (b6,h7). Black is down a rook and three pawns ($\approx$-4) but has an exposed white king on e2 and active heavy pieces (Qh1, Rh6) aimed at the white king. King safety outweighs pawn structure; if Black can win the rook on c1 (the only white piece guarding the king) the balance will swing decisively.}\\[4pt]
\end{quote}
\end{tcolorbox}
\caption{Example reasoning trace generated by OpenAI \texttt{o3} for chess position analysis, used in our reasoning SFT corpus. Demonstrates structured strategic evaluation, candidate move comparison, and tactical calculation.}
\label{figure:o3_reasoning_trace_example_1}
\end{figure}

\newpage

\begin{figure}[h]
\centering
\begin{tcolorbox}[colback=white,colframe=gray,title=OpenAI \texttt{o3} reasoning trace data example (2/2)]
\begin{quote}
\footnotesize
\texttt{Key candidate moves}\\[2pt]
\texttt{-------------------}\\
\texttt{1. Qxc1}\\[4pt]
\texttt{• Wins the rook outright, eliminating White’s only heavy defender.}\\
\texttt{• After 1…Qxc1 2.Qxb6+ (best practical try) the check is easily met by 2…Rxb6, when Black emerges a full queen vs rook up (queen + rook vs queenless side) and the attack continues.}\\
\texttt{• Other White tries (2.Qb8+, 2.Qa8+, etc.) are harmless: the black king steps to e7 or d7, and Black keeps the extra rook.}\\
\texttt{2. Qh5+}\\[4pt]
\texttt{• Initiates another wave of checks, but White can defend with 2.f3 or 2.Kd2.}\\
\texttt{• After 2.f3 Qb5+ 3.Ke1 Black still hasn’t made tangible progress and remains a rook down.}\\
\texttt{3. Qf3+}\\[4pt]
\texttt{• 2.Kd3 or 2.Ke1 hold; no forced win appears and Black is still down a rook.}\\
\texttt{4. Qe1+/Qd1+/Qf1+}\\[4pt]
\texttt{• All drop the queen to Rxc1/Rxf1.}\\

\texttt{Strategic comparison}\\[2pt]
\texttt{-------------------}\\
\texttt{Capturing on c1 converts the attack into a concrete material advantage (Black will be *ahead* in material) while leaving the white king unsafe. The attacking tries (Qh5+, Qf3+) keep pressure but rely on White going wrong; objectively they are weaker because White can consolidate and stay a rook up.}\\[4pt]

\texttt{Calculation of the critical line}\\[2pt]
\texttt{-------------------}\\
\texttt{1…Qxc1 2.Qxb6+ (the only forcing move) 2…Rxb6 – the rook on h6 slides to b6, eliminating the checking queen. Position after 2…Rxb6: Black – Qc1,Rb6,h7 ; White – Ke2, *no* queen/rook. Black is completely winning (+ material, safer king, passed b-pawn possible).}\\[4pt]

\texttt{Therefore 1...Qxc1 is the strongest move.}\\
\texttt{</think>}\\[4pt]
\texttt{<answer>Qxc1</answer>}
\end{quote}
\end{tcolorbox}
\caption{Example reasoning trace generated by OpenAI \texttt{o3} for chess position analysis, used in our reasoning SFT corpus. Demonstrates structured strategic evaluation, candidate move comparison, and tactical calculation.}
\label{figure:o3_reasoning_trace_example_2}
\end{figure}

\end{document}